\documentclass{article}

\usepackage{amsmath}
\usepackage{amssymb}
\usepackage{times}
\usepackage{bm}
\usepackage{xargs}
\usepackage{multirow}
\usepackage{amsfonts}
\usepackage{color}
\usepackage{booktabs}

\newcommandx\includeImageLineWidth[2][1=1.0]{\includegraphics[width=#1\linewidth]{#2}}

\newcommand{\bh}{{\bm{h}}}

\newcommand{\br}{{\bm{r}}}

\newcommand{\bu}{{\bm{u}}}

\newcommand{\bx}{{\bm{x}}}

\newcommand{\bz}{{\bm{z}}}

\newcommand{\bU}{{\bm{U}}}

\usepackage{array}
\newcommand{\PreserveBackslash}[1]{\let\temp=\\#1\let\\=\temp}
\newcolumntype{C}[1]{>{\PreserveBackslash\centering}p{#1}}
\newcolumntype{R}[1]{>{\PreserveBackslash\raggedleft}p{#1}}
\newcolumntype{L}[1]{>{\PreserveBackslash\raggedright}p{#1}}

\include{math_commands}

\usepackage[preprint,nonatbib]{neurips_2023}

\usepackage[utf8]{inputenc} 
\usepackage[T1]{fontenc}    
\usepackage{hyperref}       
\usepackage{url}            
\usepackage{booktabs}       
\usepackage{amsfonts}       
\usepackage{nicefrac}       
\usepackage{microtype}      
\usepackage{xcolor}         
\usepackage{algorithm}
\usepackage{algorithmic}
\usepackage{graphicx}
\usepackage{subfig}
\usepackage{subfiles}
\usepackage{enumitem}
\usepackage{wrapfig}
\usepackage{pifont}
\usepackage{wrapfig}

\definecolor{LightCyan}{rgb}{0.9059,0.9961,1}
\usepackage{multirow}
\usepackage{graphicx}
\usepackage[font=small,aboveskip=0pt,belowskip=0pt]{caption}   
\usepackage{makecell}
\usepackage{tabulary}
\definecolor{demphcolor}{RGB}{144,144,144}

\definecolor{mygray}{gray}{0.4}
\usepackage{pifont}
\newcommand{\cmark}{\color{mygray}\ding{51}}%
\newcommand{\xmark}{\color{mygray}\ding{55}}%
\newlength\savewidth

\newcommand{\tablestyle}[2]{\setlength{\tabcolsep}{#1}\renewcommand{\arraystretch}{#2}\centering\footnotesize}
\makeatletter\renewcommand\paragraph{\@startsection{paragraph}{4}{\z@}
  {.5em \@plus1ex \@minus.2ex}{-.5em}{\normalfont\normalsize\bfseries}}\makeatother
\newcommand\scalemath[1]{\scalebox{1.0}{\mbox{\ensuremath{\displaystyle #1}}}}
\usepackage{xspace}
\newcommand{\ourmodel}{SFUNet\xspace}

\usepackage{listings}

\usepackage{xcolor}

\definecolor{codegreen}{rgb}{0,0.6,0}
\definecolor{codegray}{rgb}{0.5,0.5,0.5}
\definecolor{codepurple}{rgb}{0.58,0,0.82}
\definecolor{backcolour}{rgb}{0.95,0.95,0.92}

\lstdefinestyle{mystyle}{
  backgroundcolor=\color{backcolour}, commentstyle=\color{codegreen},
  keywordstyle=\color{magenta},
  numberstyle=\tiny\color{codegray},
  stringstyle=\color{codepurple},
  basicstyle=\ttfamily\footnotesize,
  breakatwhitespace=false,         
  breaklines=true,                 
  captionpos=b,                    
  keepspaces=true,                 
  numbers=left,                    
  numbersep=5pt,                  
  showspaces=false,                
  showstringspaces=false,
  showtabs=false,                  
  tabsize=2
}

\newcommand\Mark[1]{\textsuperscript#1}

\title{Spatial-Frequency U-Net for \\ Denoising Diffusion Probabilistic Models}

\author{Xin Yuan\Mark{1}\thanks{This work has been done during the first author’s internship at Microsoft.}, Linjie Li\Mark{2}, Jianfeng Wang\Mark{2}, Zhengyuan Yang\Mark{2}, Kevin Lin\Mark{2} 
\\
\textbf{Zicheng Liu\Mark{2}, and Lijuan Wang\Mark{2}}\\
\Mark{1}University of Chicago \Mark{2}Microsoft Azure AI \\
{\tt\small yuanx@uchicago.edu
\{linjli,jianfw,zhengyang,kelin,zliu,lijuanw\}@microsoft.com}
}

\begin{document}

\maketitle

\begin{abstract}
In this paper, we study the denoising diffusion probabilistic model (DDPM) in wavelet space, instead of pixel space, for visual synthesis. Considering the wavelet transform represents the image in spatial and frequency domains, we carefully design a novel architecture \ourmodel
to effectively capture the correlation for both domains.
Specifically, in the standard denoising U-Net for pixel data, we supplement the 2D convolutions and spatial-only attention layers  with our spatial frequency-aware convolution and attention modules to jointly model the complementary information from spatial and frequency domains in wavelet data.
Our new architecture can be used as a drop-in replacement to 
the pixel-based network and is compatible with the vanilla DDPM training process.
By explicitly modeling the wavelet signals, we find our model
is able to generate images with higher quality on CIFAR-10, FFHQ, LSUN-Bedroom, and LSUN-Church datasets, than the pixel-based counterpart. 
\end{abstract}

\section{Introduction} \label{sec:intro}
The Denoising Diffusion Probabilistic Model (DDPM)~\cite{DBLP:conf/nips/HoJA20} has garnered significant attention owing to its exceptional capability to generate high-fidelity images, surpassing GANs~\cite{goodfellow2014generative,
Xu18,
Han17,Karras2019stylegan2} in
quality in many circumstances. The fundamental concept behind DDPM entails the gradual corruption of the input signal with noise, eventually conforming it to a pre-defined distribution, such as the Gaussian distribution. Subsequently, a denoising network is learned to restore the original sample, effectively removing the introduced noise.

The input signal can be from either the pixel space~\cite{DBLP:conf/icml/NicholD21,DBLP:conf/nips/DhariwalN21,DBLP:journals/corr/abs-2207-12598} or the latent space~\cite{DBLP:conf/cvpr/RombachBLEO22,DBLP:journals/corr/abs-2304-09787}. However, the focus of our investigation lies in the wavelet space. This is motivated by the widespread utilization of wavelet transforms in image processing tasks, such as image denoising~\cite{tian2023multi,huang2022winnet}. In the context of DDPM, the denoising network plays a pivotal role in eliminating noise, a task closely resembling image denoising itself. By utilizing wavelet transforms, the image can be represented in both spatial and frequency domains, facilitating explicit modeling of relations between signals across different frequencies. The goal is to leverage these characteristics to potentially achieve enhanced performance.


   
      

\begin{figure}[th]
\captionsetup[subfloat]{farskip=1pt,captionskip=1pt}
\centering
   \subfloat[\footnotesize{DDPM}]{
      \includegraphics[height = 0.07\columnwidth]{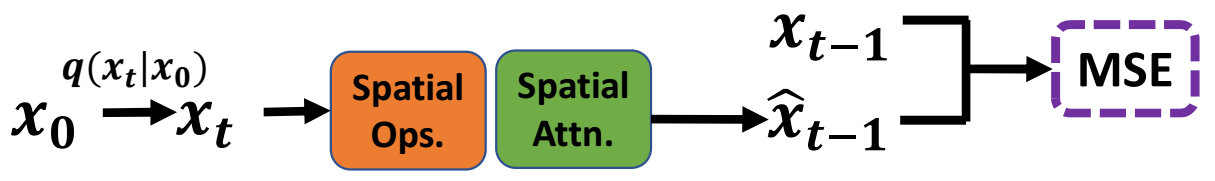}\label{fig:pixel_ddpm}
   }
   
   \subfloat[\footnotesize{WaveDiff}]{
      \includegraphics[height = 0.125\columnwidth]{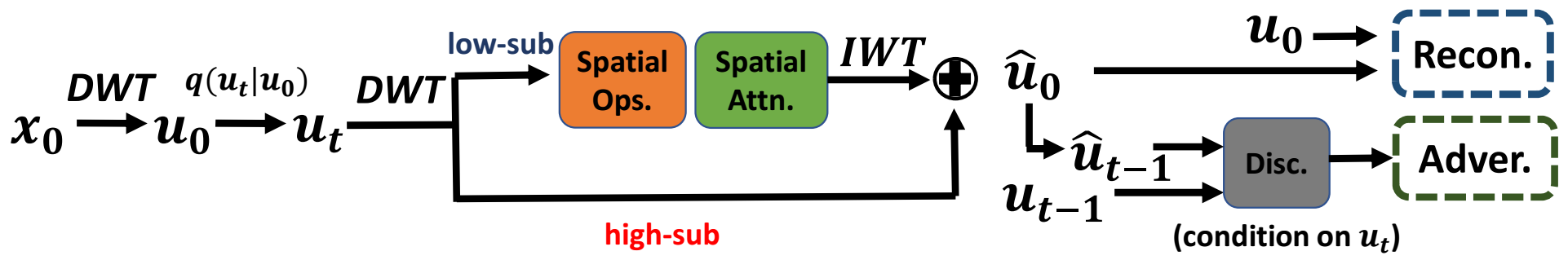}\label{fig:wavediff}
   }
   
   \subfloat[\footnotesize{\ourmodel(ours)}]{
      \includegraphics[height = 0.145\columnwidth]{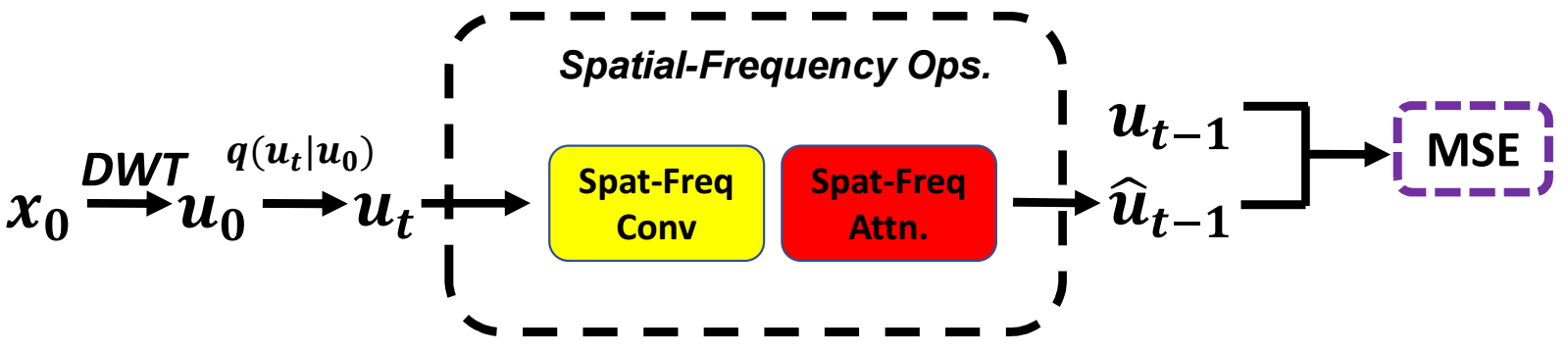}\label{fig:ours}
   }
   \caption{
     Key differences in model design among standard DDPM~\cite{DBLP:conf/nips/HoJA20}, WaveDiff~\cite{DBLP:journals/corr/abs-2211-16152} and our method. \textbf{(a)} Standard DDPM U-Net transforms noisy samples $\bx_t$ in the pixel space using spatial convolution and spatial attention to recover from the corrupted image. The optimization objective is the simple mean squared loss (MSE). \textbf{(b)} WaveDiff adopts DWT/IWT as downsample/upsample operations and only processes the low subband signal with spatial convolution and attention, followed by a inverse wavelet transform (IWT) to recover~$\bu_0$. 
     WaveDiff demands an auxiliary discriminator (Disc.), guided with reconstruction and adversarial losses. \textbf{(c)} \ourmodel, in contrast, sticks to the DDPM training process, without any extra optimization efforts. We view the samples in the wavelet space as 5D data, a combination of both low and high subbands, as an analogy to the standard DDPM U-Net, where a 4D data structure represents the pixel space. 
   }
   \vspace{-3mm}
    \label{fig:motivation}
\end{figure}

Although DDPMs in the pixel space and the latent space have been extensively studied in existing works, the exploration of the wavelet space remains relatively under-explored.
A closely related work in this field is WaveDiff~\cite{DBLP:journals/corr/abs-2211-16152}, which aims to strike a balance between efficiency and sample quality by combining Discrete/Inverse Wavelet Transformation (D/IWT) with DDGAN~\cite{DBLP:conf/iclr/XiaoKV22}. In WaveDiff, the signal undergoes wavelet transformation and inverse transformation multiple times within the network.
Meanwhile, as shown in Figure~\ref{fig:wavediff}, the training process of WaveDiff necessitates the presence of an auxiliary discriminator, guided by adversarial losses.
In contrast, our objective is to develop an effective network specifically for the wavelet space, while maintaining compatibility with the vanilla DDPM training paradigm. 
This is not a trivial task, as the computational operators (\textit{e.g.}, convolution and  attention mechanisms), and the denoising objectives are primarily tailored for pixel space diffusion.

Specifically, we propose a novel architecture \ourmodel (Figure~\ref{fig:ours}) for the diffusion and denoising process purely in wavelet space. With the wavelet transform (Haar transform~\cite{DBLP:journals/siamrev/Brewster93} in our experiments), the 2D input image is mapped into a 3D signal, where two dimensions represent the spatial domains and one represents frequency. To fully correlate the frequency domain, we incorporate a 1D convolutional layer along the frequency dimension in addition to the 2D convolutional layers along the spatial dimensions.
Recognizing that convolution is less effective at capturing global correlations, we further enhance our model by incorporating attention mechanisms. We apply attention at both each spatial location across different frequencies and each frequency across different spatial locations. 
By employing these separable modules, we can efficiently process high-dimensional inputs, offering improved performance compared to full 3D convolutional layers and full attention mechanisms that operate on all locations and frequencies.

During training, we optimize the model using the mean squared error (MSE) loss to predict noise over all diffusion timesteps, following the vanilla diffusion model in pixel space. During inference, the restored wavelet signal is transformed back into the pixel space, leveraging the reversibility of the wavelet transform. Despite its simplicity, our model consistently outperforms existing approaches on multiple datasets, including CIFAR-10~\cite{cifar10}, FFHQ-256~\cite{DBLP:conf/cvpr/KarrasLA19}, LSUN-Bedroom-256~\cite{LSUN}, and LSUN-Church-256~\cite{LSUN}.

We summarize our contributions as two-fold:
\vspace{-0.5em}
\begin{itemize}[leftmargin=.15in]
\setlength\itemsep{0.2em}
    \item{%
      \textbf{Spatial-Frequency-aware Architectural Design.}~
      The architecture of \ourmodel is specifically and carefully designed for wavelet data. By explicitly processing and exploiting the information from both spatial and frequency subspaces, the distribution of image contents (\textit{i.e.,} spatial components across different frequencies)  and local details (\textit{i.e.,} high-frequency components) can be better converged to reverse the forward diffusion process. The newly designed modules in \ourmodel can be easily dropped in to DDPM U-Net without affecting the default structure.
   }%
   \item{%
      \textbf{High Quality of Image Generation.}~
      Our \ourmodel can generate high-quality images with clear details, achieving excellent quantitative and qualitative results under common evaluation protocols. 
   }%
\end{itemize}
\section{Related Work}
The denoising U-Net~\cite{DBLP:conf/nips/HoJA20,DBLP:conf/miccai/RonnebergerFB15} is an essential design to the success of the DDPMs in generation tasks~\cite{DBLP:conf/cvpr/RombachBLEO22,DBLP:conf/icml/NicholD21,DBLP:conf/nips/DhariwalN21,DBLP:journals/corr/abs-2207-12598,DBLP:conf/icml/NicholDRSMMSC22}.
As important building blocks, 2D convolutions~\cite{DBLP:conf/cvpr/HeZRS16} and spatial self-attention~\cite{DBLP:conf/nips/VaswaniSPUJGKP17} effectively extract intermediate features from images, proven successful for the denoising task.
Video diffusion models~\cite{DBLP:conf/nips/HoSGC0F22,DBLP:journals/corr/abs-2209-14792,DBLP:journals/corr/abs-2210-02303}, inspired by video understanding models 
~\cite{DBLP:conf/iccv/TranBFTP15,DBLP:conf/cvpr/CarreiraZ17,DBLP:conf/cvpr/TranWTRLP18,li2019global,li2019multi,chen2020temporal},
propose a new type of U-Net architecture following the principle of jointly extracting spatial-temporal information from video frames.
Such developments aim to bridge the gap between tasks (\textit{e.g.}, from image to video) with a minimum design effort in optimization recipe, in which the domain/modality shift of the data may pose a non-trivial task to the model training if the network architecture remain unchanged.

Wavelet-based deep learning approaches~\cite{DBLP:conf/icmla/WilliamsL16,DBLP:conf/eccv/YaoPLNM22,DBLP:journals/corr/abs-2102-06108} have shown great potential in providing inherent advantages that are not available in the pixel space, inspiring researchers to 
incorporate wavelets in diffusion models for generation tasks.
WaveDiff~\cite{DBLP:journals/corr/abs-2211-16152} builds upon a GAN-based method, DDGAN~\cite{DBLP:conf/iclr/XiaoKV22}, and incorporates wavelet transformation from the perspective of image compression, hence achieving a better trade-off between efficiency and sample quality.
DiWa~\cite{DBLP:journals/corr/abs-2304-01994} combines wavelets and diffusion model to improve image super-resolution by leveraging the power of high-frequency information for detail enhancement.
~\cite{DBLP:journals/corr/abs-2302-00190} proposes a diffusion model on a continuous implicit representation in wavelet space for 3D shape generation.
Among them, WaveDiff is perhaps the most relevant study to ours.
However, one important difference in design principle between our \ourmodel and WaveDiff is that: \ourmodel does not perform discrete/inverse wavelet transform (D/IWT) within each computational block. We note that, it may not be an appropriate practice to embed DWT and IWT to intermediate features, in which case, DWT and IWT just serves as differentiable linear operators.  
Moreover, with GAN~\cite{goodfellow2014generative,DBLP:conf/iclr/XiaoKV22} components embedded in WaveDiff, including the auxiliary discriminator and the adversarial training objectives, one may not have the flexibility to further explore whether the noisy wavelet signals can be effectively recovered only through the reversed diffusion process, which is the question we aim to answer in this paper.

\section{Method}
Given an input image $\scalemath{\bx_0}$ of resolution $\scalemath{H \times W}$, we utilize the discrete wavelet transform (DWT) (Haar wavelet~\cite{DBLP:journals/siamrev/Brewster93}) to decompose it into four subbands. 
We define the low-pass filter $L=\frac{1}{\sqrt{2}}[1,1]^T$ and the high-pass filter $H=\frac{1}{\sqrt{2}}[-1,1]^T$. 
With these filters, we construct four convolutional kernels: $LL^T$, $LH^T$, $HL^T$, and $HH^T$. 
The $LL^T$ kernel effectively performs average pooling over $2\times 2$ windows, capturing low-frequency components. 
The other kernels extract different higher-level frequencies. 
By applying these kernels, we represent the input image  $\scalemath{\bx_0}$  as 
$\scalemath{\bu_0 = [\bU_{ll}, \bU_{lh}, \bU_{hl}, \bU_{hh}]}$ ($\bU_{*} \in \mathcal{R}^{H/2 \times W/2}$).
Note that, each subband has a downsampled resolution with a factor of 2 compared to the original image.
Importantly, this decomposition is exactly invertible using the inverse wavelet transform (IWT), preserving all the information in the pixel space.
To this end, the input image $\scalemath{H \times W}$ is transformed to a 3D signal $\scalemath{4 \times H/2 \times W/2}$.
Instead of concatenating the subbands along the channel dimension~\cite{DBLP:journals/corr/abs-2211-16152,DBLP:journals/corr/abs-2304-01994}, 
we adopt, in practice, a 5D data structure of $\scalemath{[B, C, F, H/2, W/2]}$, where $\scalemath{B}$, $\scalemath{C}$ and $\scalemath{F}$ denote the 
batch size, number of channels and subbands, respectively. 
Initially, the channel $C$ is set to 3, representing the R, G, and B channels.
We explicitly separate feature, frequency and spatial dimensions to facilitate the parallel processing of different wavelet subspaces within our proposed spatial-frequency block. 

As an analogy to the pixel DDPM~\cite{DBLP:conf/nips/HoJA20}, we generate a noisy wavelet sample $\scalemath{\bu_t}$ from $\scalemath{\bu_0}$ through the forward diffusion in the wavelet space:
\begin{eqnarray}\label{eq:q_sample}
    \scalemath{
    q(\bu_t|\bu_0):= \mathcal{N}(\bu_t; \sqrt{\bar{\alpha}_t}\bu_0, (1-\bar{\alpha}_t)I),
    } \nonumber \\
    \scalemath{
    \bu_{t} = \sqrt{\bar{\alpha}_t}\bu_0 + \sqrt{1-\bar{\alpha}_t}\epsilon, \epsilon \sim \mathcal{N}(0,1),
    }
\end{eqnarray}
where $\scalemath{\alpha_t = 1 - \beta_t, \bar{\alpha}_t = \prod_{s=1}^t \alpha_t}$.
Next, we first describe our architecture in Sec.~\ref{sec:unet} to recover the noise from this noisy $u_t$, and then details the training and inference in Sec.~\ref{sec:train_infer}.

\begin{figure}[h]
\begin{center}
\centerline{\includegraphics[width=\columnwidth]{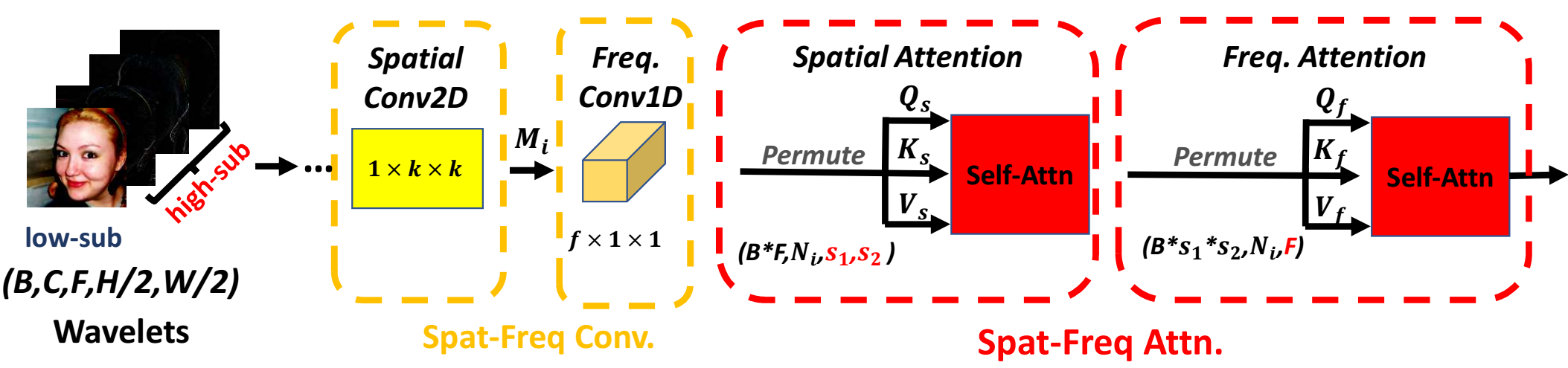}}
\caption{Spatial-frequency U-Net block: The proposed denoising U-Net block consists of two components: spatial-frequency convolution and spatial-frequency attention.  
The spatial-frequency convolution `upgrades' 2D convolution by decomposing the full 3D convolution into a spatial 2D convolution and a frequency 1D convolution to facilitate feature extraction in 5D wavelet data.
The spatial-frequency attention mechanism exploits the contribution at each spatial and frequency location. With the appropriate permutation operations, features in wavelet space can be easily adapted to attention layer without alternating the operations in self-attention layer.
The proposed spatial-frequency convolution and attention layers serve as new building blocks in the denoising U-Net encoder-decoder architecture, optimized by the simple DDPM objective (\textit{i.e.}, MSE loss).
}
\label{fig:framework}
\end{center}
\vspace{-6mm}
\end{figure}

\subsection{Spatial-Frequency U-Net}\label{sec:unet}
To accomplish the denoising task in the wavelet space while maintaining the compatibility to the standard DDPM training process, we propose two spatial-frequency components to serve as an replacement of the original design in DDPM U-Net.
Figure~\ref{fig:framework} shows the overview of the proposed spatial-frequency U-Net block. 

\noindent \textbf{Spatial-Frequency Convolution}
Suppose we have the wavelet 5D input data of size $\scalemath{[B\times N_{i-1} \times F \times s_1 \times s_2]}$ for $i$-th block, where $N_{i - 1}$ is the channel size.
One 
straightforward
design is to upgrade 2D convolutional filters of size $\scalemath{N_{i-1} \times k\times k}$ to full 3D convolutions, in which each filter is 4-dimensional ($\scalemath{N_{i-1} \times f \times k\times k}$) ($f$ is the kernel size for frequency) and all $\scalemath{N_i}$ filters are convolved over spatial and frequency dimension together.  

For computational efficiency, we frame a (2+1)D convolution, built upon~\cite{DBLP:conf/cvpr/TranWTRLP18} within the context of spatial-frequency domain: similar to the spatial-temporal modeling, the spatial-frequency convolution consists of a 2D convolution followed by a 1D convolution, which are capable of approximating the full 3D convolution while allowing joint learning capability over both spatial and frequency subspaces to emerge as needed.
As shown in Figure~\ref{fig:framework}, 
The new convolution module consists of $\scalemath{M_i}$ 2D filters of size $\scalemath{N_{i-1} \times 1 \times k \times k}$ and $\scalemath{N_i}$ 1D filters of size $\scalemath{M_i \times f \times 1 \times 1}$, in which $\scalemath{M_i}$ is $[\frac{fk^2N_{i-1}N_{i}}{k^2n_{i-1}+fN_i}]$ to approximate the parameters of a full 3D convolution.

\noindent \textbf{Spatial-Frequency Attention}
Attention layers in WaveDiff U-Net ignore the frequency ordering
in the wavelet space and process different subbands analogously to channels. 
Considering the complementary information among low-sub and high-sub(s), 
we design a simple yet effective attention mechanism with both spatial and frequency.
We build both attention layers upon the scaled dot attention\cite{DBLP:conf/nips/VaswaniSPUJGKP17}: 
\vspace{-0.5em}
\begin{eqnarray}\label{eq:attention}
    \scalemath{\text{Attention}(Q,K,V) = \bm{m} \cdot V, }\nonumber \\
    \scalemath{\bm{m} = \text{Softmax}(\frac{QK^T}{\sqrt{d}})}
\end{eqnarray}
where $d$ is the dimension of queries and keys.
In the context of self-attention, individual entries of the mask $\bm{m}_{i,j}$ represent the contribution of the $j$-th location towards the $i$-th one.
Given intermediate features processed by the spatial-frequency convolutions, denoted as $\scalemath{\bh_{u}}$ of shape $\scalemath{B\times N_i \times F \times s_1 \times s_2}$, we have the flexibility to permute the data structure to realize both spatial and frequency attention while maintaining the unified definition in Eq.~\ref{eq:attention}. 
Specifically, for spatial attention, we permute $\scalemath{\bh_{u}}$ to be the shape of $\scalemath{(B*F , N_i , s_1*s_2)}$, denoted as $\hat{\bh}_{u}$. Then Query $\scalemath{Q_s = W^{Q}_s(\hat{\bh}_{u})}$, Key $\scalemath{K_s = W^{K}_s(\hat{\bh}_{u})}$, and Value $\scalemath{V_s = W^{V}_s(\hat{\bh}_{u})}$ are computed with the learnable projections $\scalemath{W^{Q}_s}$, $\scalemath{W^{K}_s}$ and $\scalemath{W^{V}_s}$ applied on $\scalemath{\hat{\bh}_{u}}$ and processed with Eq.~\ref{eq:attention} to generate $\scalemath{\hat{\br}_{u}}$.
Similarly for frequency attention, we permute $\scalemath{\br_{u}}$ to be the shape of $\scalemath{(B*s_1*s_2, N_i,F)}$, compute $\scalemath{Q_f}$, $\scalemath{K_f}$ and $\scalemath{V_f}$ using learnable projections, which is followed by self-attention defined in Eq.~\ref{eq:attention}.

We construct our spatial-frequency U-Net, parameterized as $\theta$, with the proposed spatial-frequency convolution and attention blocks, and make the noise prediction $\hat{\epsilon}_{\theta}$ of shape $\scalemath{B \times 3 \times F \times H/2 \times  W/2}$.

\subsection{Training and Sampling with Standard Denoising Diffusion}\label{sec:train_infer}
We train the model in an end-to-end manner, with the simple denoising objective in DDPM.
As such, the model weights $\theta$ can be optimized by minimizing the MSE loss of noise prediction:
\begin{eqnarray} \label{eq:loss}
    \scalemath{L = \mathbb{E}||\epsilon - \hat{\epsilon_{\theta}}||_2^2}
\end{eqnarray}
We summarize the training in Algorithm~\ref{alg:training}.

After the training finishes, we sample realistic images with reversed diffusion process, starting from the input noise $\scalemath{\bu_T \sim \mathcal{N}(0,1)}$ of shape $\scalemath{B \times 3 \times F \times H/2 \times  W/2}$.
The reversed diffusion process is to predict $\bu_{t-1}$ from $\bu_{t}$, which is formulated as:
\begin{eqnarray}
    \scalemath{\bu_{t-1} = \frac{1}{\sqrt{\alpha_t}}(\bu_t -\frac{1-\alpha_t}{\sqrt{1-\bar{\alpha_t}}}\theta(\bu_t,t))+\sigma_t\bz,} \label{eq:r_diff}\\
    \scalemath{\bz \sim \mathcal{N}(0,1) \quad \text{if} \quad  t>1 \quad  \text{else} \quad  \bz=0.} \label{eq:sample_z}
\end{eqnarray}
where  $\scalemath{\sigma_t}$ is empirically set according to the noise scheduler~\cite{DBLP:conf/nips/HoJA20}. Our model performs T steps of the reversed diffusion process to produce the generation of all frequency components in the wavelet space. We then reconstruct the image $\scalemath{\hat{\bx}_0}$ using inverse wavelet transform (IWT) when $\scalemath{t=1}$. We summarize this sampling process in Algorithm~\ref{alg:eval}.

\begin{figure}[tp]
\begin{minipage}{0.48\columnwidth}
\begin{algorithm}[H]
\caption{: Training}
\label{alg:training}
\begin{algorithmic}
  \STATE {\bfseries Input:} Data $\scalemath{\bm{x}_0}$.
  \STATE {\bfseries Output:} Trained model $\scalemath{\theta}$
\STATE Initialize: Model weights $\scalemath{\theta}$, Timesteps T.
  \FOR{$\text{iter}\scalemath{=1}$ {\bfseries to }Iter$_{total}$}
      \STATE Sample  $\scalemath{t \in [1,T]}$
      \STATE Transform $\scalemath{\bm{x}_0}$ to $\scalemath{\bm{u}_0}$ using DWT.
      \STATE Sample $\scalemath{\bm{u}_t}$ using Eq.~\ref{eq:q_sample} 
      \STATE Generate $\scalemath{\hat{\epsilon}_{\theta}}$ using $\scalemath{\theta}$  
      \STATE Back propagation with Eq.~\ref{eq:loss}.
      \STATE Update $\scalemath{\theta}$.
  \ENDFOR
  \STATE return $\scalemath{\theta}$
\end{algorithmic}
\end{algorithm}
\end{minipage}
\hfill
\begin{minipage}{0.48\columnwidth}
\begin{algorithm}[H]
\caption{: Sampling}
\label{alg:eval}
\begin{algorithmic}
    \STATE {\bfseries Input:} Noise $\scalemath{\bu_T}$, trained model  $\scalemath{\theta}$.
  \STATE {\bfseries Output:} Image $\scalemath{\hat{\bx_0}}$.
\STATE Initialize: $\scalemath{\bu_T \sim \mathcal{N}(0,1)}$
  \FOR{$\text{t}=T$ {\bfseries to } 1}
      \STATE Sample  $\bz$ using Eq.~\ref{eq:sample_z}
      \STATE Perform reversed diffusion to get $\scalemath{\hat{\bu}_{t-1}}$ using Eq.~\ref{eq:r_diff}
      \IF{$\scalemath{t = 1}$}
      \STATE Transform $\scalemath{\hat{\bu}_{0}}$ to $\scalemath{\hat{\bx_0}}$ using IWT.
      \STATE return $\scalemath{\hat{\bx_0}}$.
     \ENDIF
  \ENDFOR
\end{algorithmic}
\end{algorithm}
\end{minipage}
\vspace{-5mm}
\end{figure}

\section{Experiments}

\subsection{Experimental Setup}
We evaluate our model on four datasets: CIFAR-10~\cite{cifar10}, FFHQ-256~\cite{DBLP:conf/cvpr/KarrasLA19}, LSUN-Bedroom-256~\cite{LSUN} and LSUN-Church-256~\cite{LSUN}.
Following the standard practice in existing works, we generate 50,000 images on each dataset randomly for evaluation.
We compare the model performance in the generation quality by reporting Fr{\'{e}}chet Inception Distance (FID)~\cite{DBLP:conf/nips/HeuselRUNH17}. 
We also report Precision (Prec.) and Recall (Rec.) metrics~\cite{DBLP:conf/nips/KynkaanniemiKLL19} to separately measure the sample fidelity and diversity.

We train our model on CIFAR-10 at $32\times 32$ resolution, and FFHQ, LSUN-Bedroom, LSUN-Church on $256\times 256$ resolution.
For all of our experiments, similar to~\cite{DBLP:conf/nips/HoJA20}, we use the encoder-middle-decoder architecture to construct the U-Net with our proposed convolution and attention layers.
For $32 \times 32$ image resolution, we detail the architecture as follows. The downsampling block is 4-step, each with 3 residual blocks. The upsampling block mirrors the downsampling one. 
From highest to lowest resolution, the U-Net stages adopt the channel size of $[c,2c,2c,2c]$, respectively.
We use four attention heads at the $16\times 16$ and $8\times 8$ resolution.
For model architecture with $256 \times 256$ image resolution, the down/up-sampling block is 6-step with channel sizes of $[c,c,2c,2c,4c,4c]$, and 2 residual blocks for each step, respectively. We use a single attention head at the $16\times 16$ resolution. $C$ is set as 128 for all models.

We use Adam~\cite{kingma2014adam} optimizer to train all models with a learning rate of $10^{-4}$ and an exponential moving average (EMA) over model parameters with rate $0.9999$. We adopt the linear noise scheduler in~\cite{DBLP:conf/nips/HoJA20} with $T=1000$ timesteps. 
Our CIFAR-10 model is trained on 8 Nvidia V100 32GB GPUS for 500K iterations, with a batch size of 128 and dropout of 0.1.
To accommodate for larger resolution in  FFHQ, LSUN-Bedroom and LSUN-Church, we train our models on 32 GPUS for 250K iterations.


\begin{table}[tbh]
\begin{center}
\vspace{-4mm}
\tablestyle{3.5pt}{1.0} 
\caption{Results on CIFAR-10 dataset.}
\label{tab:c10:result}
\begin{tabular}
{lccccccccc}
\toprule
&\multicolumn{1}{c}{{Architecture}}
&\multicolumn{1}{c}{{Model Type}}
&\multicolumn{1}{c}{{Denoising Space}}
&\multicolumn{1}{c}{FID$ \downarrow$}
&\multicolumn{1}{c}{Perc. $\uparrow$}
&\multicolumn{1}{c}{Rec. $\uparrow$} \\
\midrule
&DDGAN~\cite{DBLP:conf/iclr/XiaoKV22} &Diffusion + GAN &Pixel &3.75(-0.00) &- &0.57(+0.00)  \\
&WaveDiff~\cite{DBLP:journals/corr/abs-2211-16152} &Diffusion + GAN &Wavelet &4.01(+0.26) &- &0.55(-0.02) \\
\midrule
&DDPM~\cite{DBLP:conf/nips/HoJA20} U-Net (default) &Diffusion &Pixel &3.53(-0.00) &0.62(+0.00) &0.55(+0.00) \\
&DDPM~\cite{DBLP:conf/nips/HoJA20} U-Net (concat) &Diffusion &Wavelet &9.29(+5.76) &0.64(+0.02) &0.51(-0.04)\\
&\ourmodel (ours) &Diffusion &Wavelet &4.88(+1.35) &0.60(-0.02) &0.55(+0.00)\\
\bottomrule
\end{tabular}
\vspace{-5mm}
\end{center}
\end{table}

\subsection{Quantitative Results}
We report generation quality in terms of FID, Precision and Recall, and compare with several baselines. With the same denoising objective and training process, the default DDPM U-Net architecture in pixel space is a natural and trivial baseline. We also design a simple variant of the denoising U-Net to process noisy wavelet inputs in a 4D form with all subbands concatenated along the channel dimension. Such data structure results in a corresponding change in the number of first/last convolution's input/output channels. (i.e. from 3 to 12). We denote this baseline as DDPM U-Net (concat). On CIFAR-10 and LSUN-Church-256, we additionally compare with the results reported in DDGAN~\cite{DBLP:conf/iclr/XiaoKV22} and WaveDiff~\cite{DBLP:journals/corr/abs-2211-16152}. Note that the DDPM U-Net (concat) is a simplified version of WaveDiff, removing the excessive D/IWT operators and the discriminator for GAN training.

As shown in Table~\ref{tab:c10:result}, though \ourmodel and WaveDiff can achieve plausible performance on CIFAR-10, both of them cannot beat the corresponding pixel-space counterparts (\textit{i.e.}, DDPM U-Net (default) and DDGAN). We hypothesize that this is due to the low image resolution in CIFAR-10, where transforming from pixel space to wavelet space further lowers the input resolution by half, thereby can not provide much benefit. \ourmodel also achieves comparable performance to WaveDiff on CIFAR-10, it is worth noting that \ourmodel adopts a simpler architectural design and training objective than WaveDiff, which embeds DWT and IWT as differentiable operators multiple times inside the network, and is trained with adversarial loss against an auxiliary discriminator.
Furthermore, the `concat' baseline yields a poorer generation performance than \ourmodel, which suggests naively modifying the input channel to the standard U-Net cannot facilitate the denoising process in wavelet space. 

Table~\ref{tab:ffhq:result} shows a significant improvement of FID from \ourmodel over baselines on FFHQ-256.
These results deliver several findings, (1) the benefit of directly denoising in wavelet space is more substantial for higher resolution image generation; and (2) \ourmodel is able to fully exploit the information from wavelet space hence producing more realistic face images, with much lower FID.
Similar conclusions can be drawn from Table~\ref{tab:church:result} and ~\ref{tab:bedroom:result} on LSUN datasets, \ourmodel consistently improves over the pixel-space counterpart and the `concat' baseline, which validates the proposed architecture in \ourmodel  can successfully model the wavelets information within the context of standard DDPM training.

\begin{table}[tbh]
\begin{center}
\vspace{-4mm}
\tablestyle{4pt}{1.0} 
\caption{Results on FFHQ-256 dataset.}
\label{tab:ffhq:result}
\begin{tabular}
{lccccccccc}
\toprule
&\multicolumn{1}{c}{{Architecture}}
&\multicolumn{1}{c}{{Model Type}}
&\multicolumn{1}{c}{{Denoising Space}}
&\multicolumn{1}{c}{FID$ \downarrow$}
&\multicolumn{1}{c}{Perc. $\uparrow$}
&\multicolumn{1}{c}{Rec. $\uparrow$} \\
\midrule
&DDPM~\cite{DBLP:conf/nips/HoJA20} U-Net (default) &Diffusion &Pixel &13.53(-0.00) &0.52(+0.00) &0.31(+0.00) \\
&DDPM~\cite{DBLP:conf/nips/HoJA20} U-Net (concat) &Diffusion &Wavelet &23.18(+9.65) &0.55(+0.03) &0.30(-0.01)\\
&\ourmodel (ours) &Diffusion &Wavelet &7.12(-6.41) &0.54(+0.02) &0.38(+0.07)\\
\bottomrule
\end{tabular}
\end{center}
\vspace{-8mm}
\end{table}

\begin{table}[htb]
\begin{center}
\tablestyle{5pt}{1.0} 
\caption{Results on LSUN-Church-256 dataset.}
\label{tab:church:result}
\begin{tabular}
{lccccccccc}
\toprule
&\multicolumn{1}{c}{{Architecture}}
&\multicolumn{1}{c}{{Model Type}}
&\multicolumn{1}{c}{{Denoising Space}}
&\multicolumn{1}{c}{FID$ \downarrow$}
&\multicolumn{1}{c}{Perc. $\uparrow$}
&\multicolumn{1}{c}{Rec. $\uparrow$} \\
\midrule
&DDGAN~\cite{DBLP:conf/iclr/XiaoKV22} &Diffusion + GAN &Pixel &5.25(-0.00) &- &-\\
&WaveDiff~\cite{DBLP:journals/corr/abs-2211-16152} &Diffusion +GAN &Wavelet &5.06(-0.19) &- &0.40\\
\midrule
&DDPM~\cite{DBLP:conf/nips/HoJA20} U-Net (default) &Diffusion &Pixel &7.89(-0.00) &- &-  \\
&DDPM~\cite{DBLP:conf/nips/HoJA20} U-Net (concat) &Diffusion &Wavelet &18.96(+11.07) &0.61 &0.41 \\
&\ourmodel (ours) &Diffusion &Wavelet &6.11(-1.78) &0.60 &0.44\\
\bottomrule
\end{tabular}
\vspace{-6mm}
\end{center}
\end{table}

\begin{table}[htb]
\begin{center}
\tablestyle{4pt}{1.0} 
\caption{Results on LSUN-Bedroom-256 dataset.}
\label{tab:bedroom:result}
\begin{tabular}
{lccccccccc}
\toprule
&\multicolumn{1}{c}{{Architecture}}
&\multicolumn{1}{c}{{Model Type}}
&\multicolumn{1}{c}{{Denoising Space}}
&\multicolumn{1}{c}{FID$ \downarrow$}
&\multicolumn{1}{c}{Perc. $\uparrow$}
&\multicolumn{1}{c}{Rec. $\uparrow$} \\
\midrule
&DDPM~\cite{DBLP:conf/nips/HoJA20} U-Net (default) &Diffusion &Pixel &4.89(-0.00) &0.60(+0.00) &0.45(+0.00) \\
&DDPM~\cite{DBLP:conf/nips/HoJA20} U-Net (concat) &Diffusion &Wavelet &18.23(+13.34) &0.44(-0.16) &0.41(-0.04) \\
&\ourmodel (ours) &Diffusion &Wavelet &3.88(-1.01) &0.62(-0.02) &0.48(+0.03) \\
\bottomrule
\end{tabular}
\vspace{-4mm}
\end{center}
\end{table}

\begin{figure}[tbh]
\captionsetup[subfloat]{farskip=1pt,captionskip=1pt}
 \begin{center}
   \subfloat[Generated Wavelets]{
      \includegraphics[height = 0.48 \columnwidth]{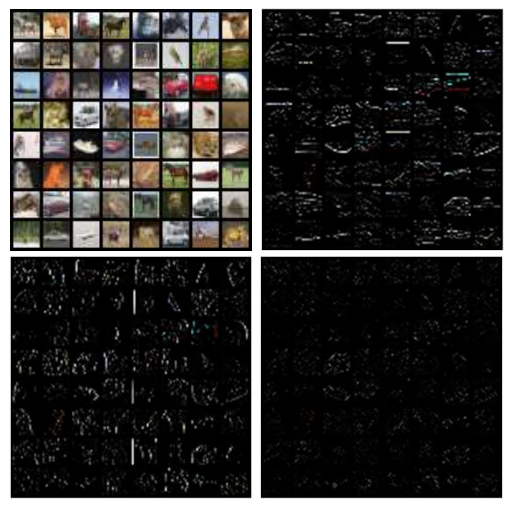}\label{fig:cifar_a}
   }\hfill
   \subfloat[Generated Images ]{
      \includegraphics[height = 0.48 \columnwidth]{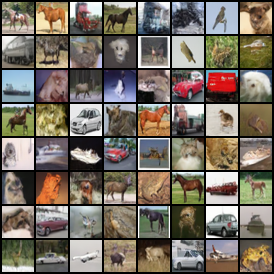}\label{fig:cifar_b}
   }
\end{center}
   \caption{
     Visualization of generations on CIFAR-10 (\textit{best viewed when zoomed-in, especially for generated high-frequency wavelets}). 
   }
   \vspace{-4mm}
   \label{fig:c10}
\end{figure}

\begin{figure}[tbh]

\captionsetup[subfloat]{farskip=1pt,captionskip=1pt}
 \begin{center}
 \subfloat[Generated Wavelets ]{
      \includegraphics[height = 0.48 \columnwidth]{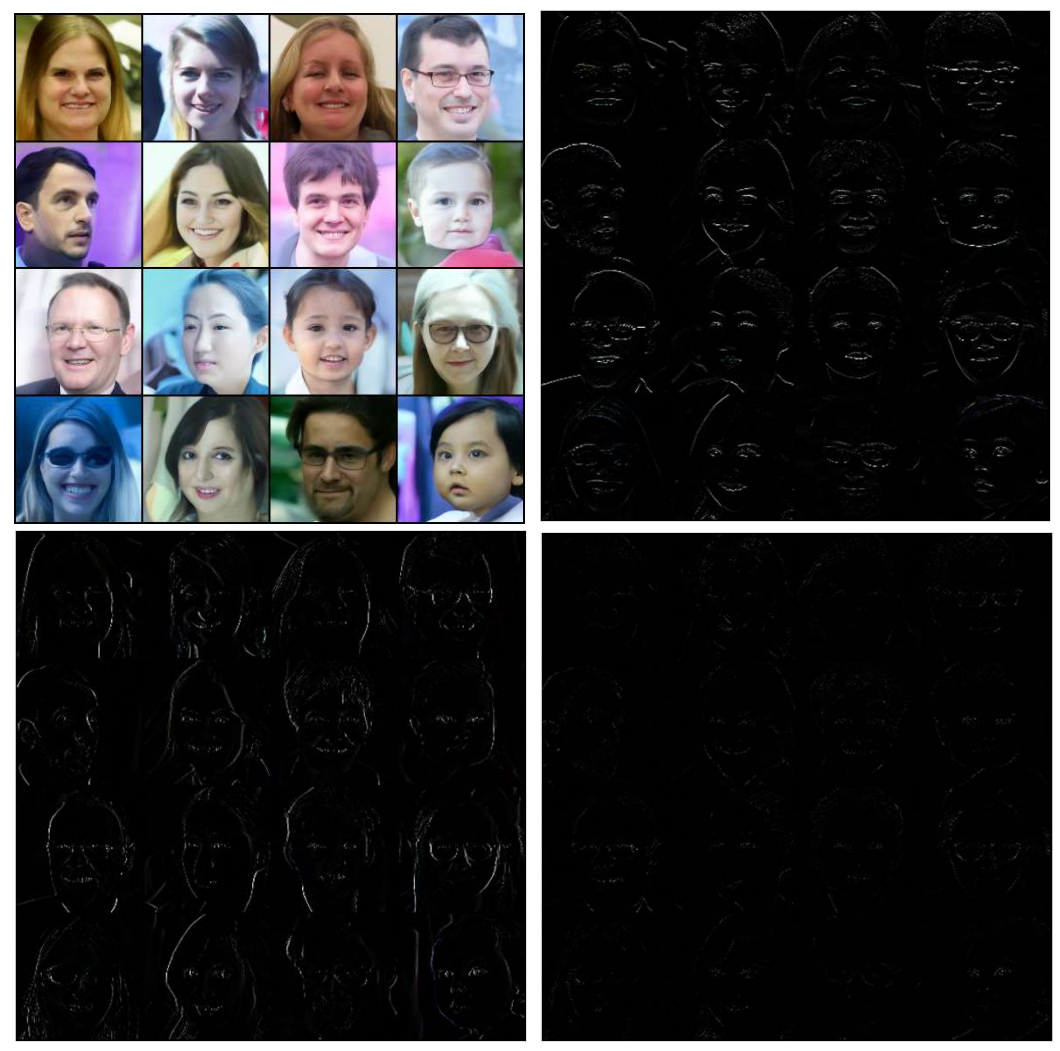}\label{fig:ffhq_a}
   }\hfill
   \subfloat[Generated Images]{
      \includegraphics[height = 0.48 \columnwidth]{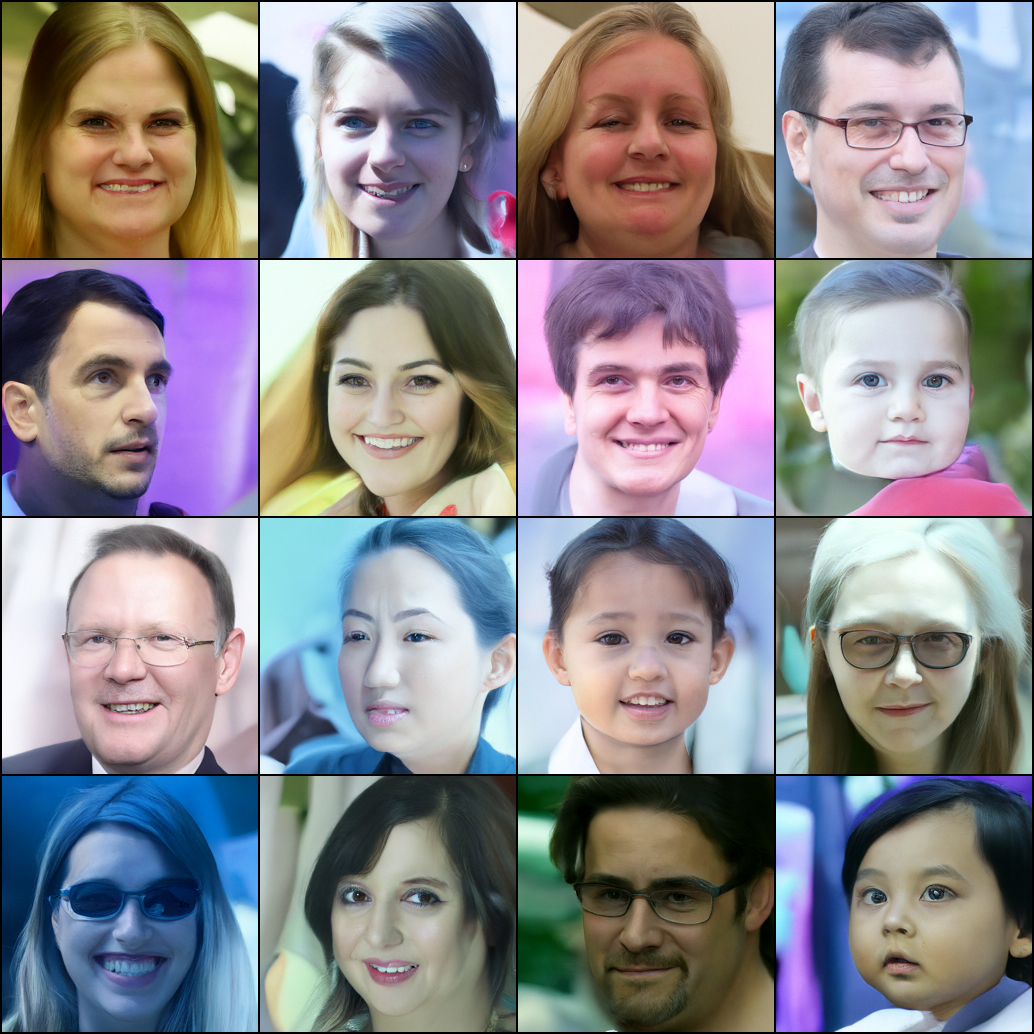}\label{fig:ffhq_b}
   }
\end{center}
   \caption{
     Visualization of generations on FFHQ-256 (\textit{best viewed when zoomed-in}).
   }
   \vspace{-4mm}
   \label{fig:ffhq}
\end{figure}

\begin{figure}[tb]
\captionsetup[subfloat]{farskip=1pt,captionskip=1pt}
 \begin{center}
   \subfloat[Generated Wavelets]{
      \includegraphics[height = 0.48\columnwidth]{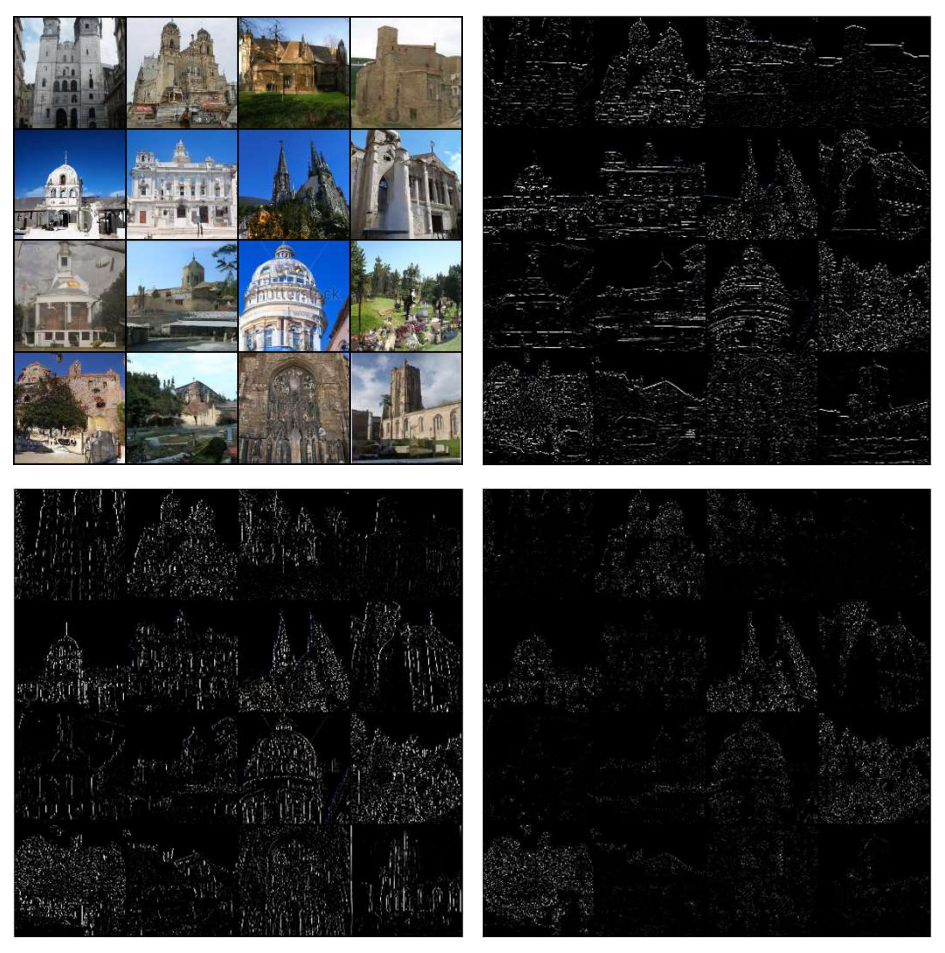}\label{fig:church_a}
   }\hfill
   \subfloat[Generated Images ]{
      \includegraphics[height = 0.48 \columnwidth]{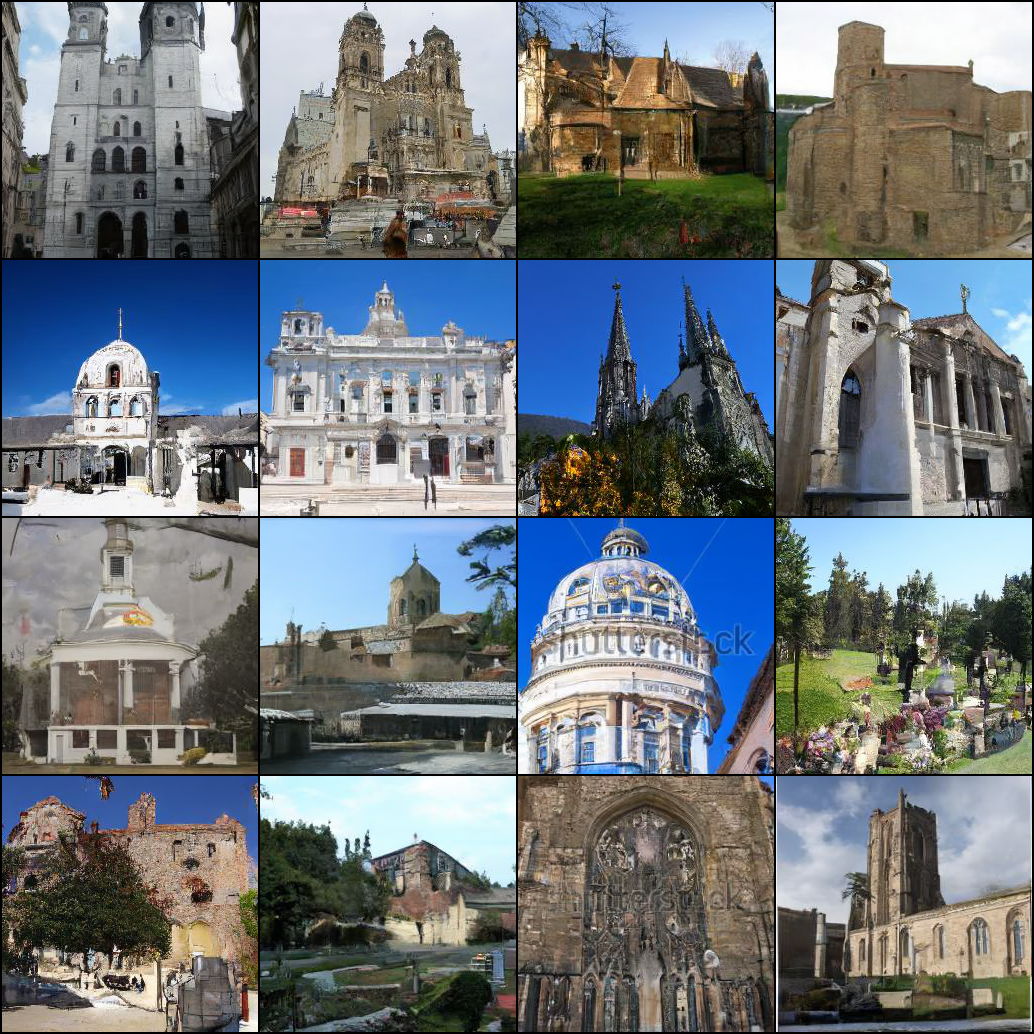}\label{fig:church_b}
   }
\end{center}
   \caption{
     Visualization of generations on LSUN-Church-256 (\textit{best viewed when zoomed-in}).
   }
   \vspace{-4mm}
   \label{fig:church}
\end{figure}

\begin{figure}[tb]
\captionsetup[subfloat]{farskip=1pt,captionskip=1pt}
 \begin{center}
   \subfloat[Generated Wavelets]{
      \includegraphics[height = 0.48 \columnwidth]{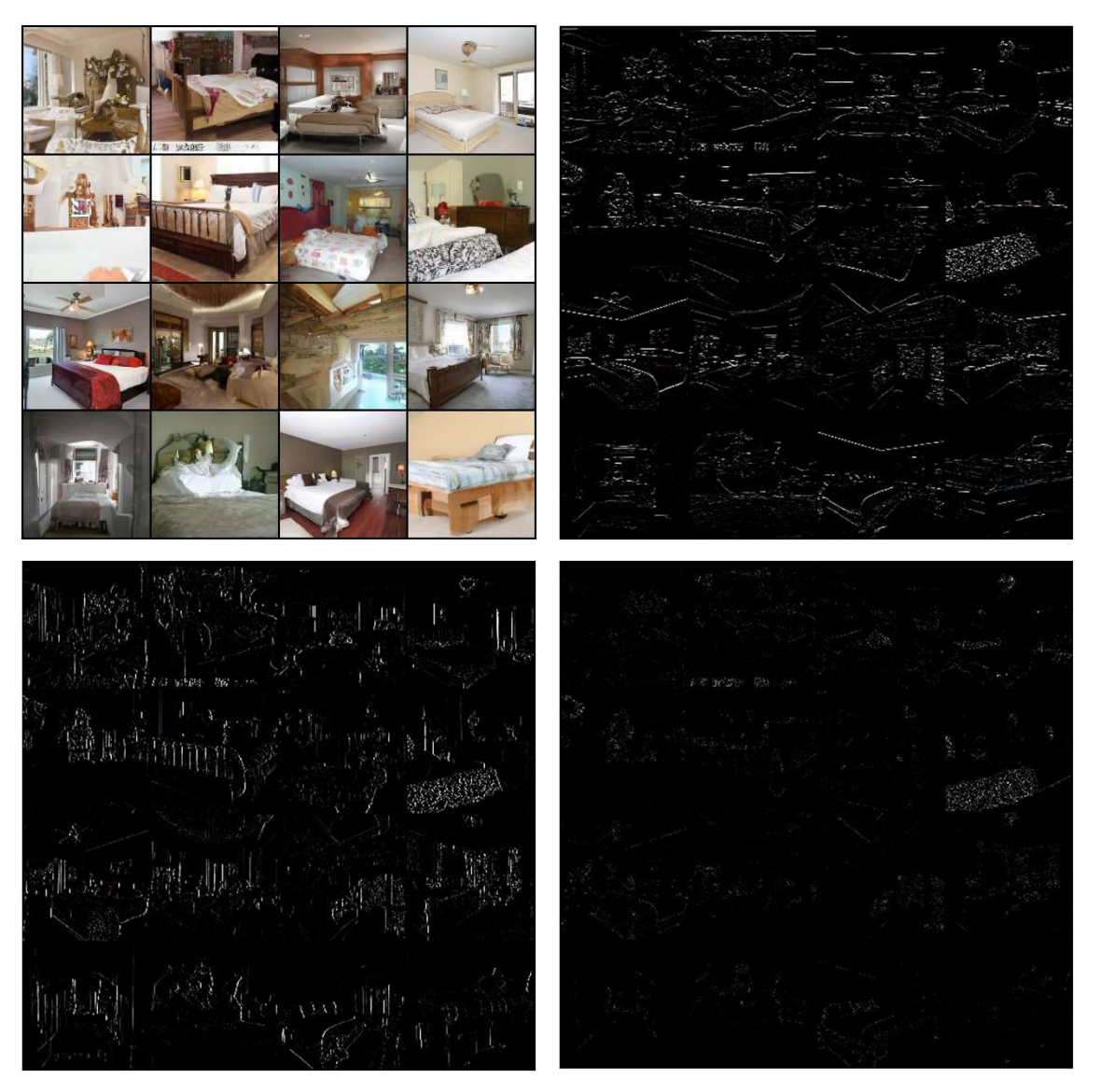}\label{fig:bedroom_a}
   }\hfill
   \subfloat[Generated Images]{
      \includegraphics[height = 0.48 \columnwidth]{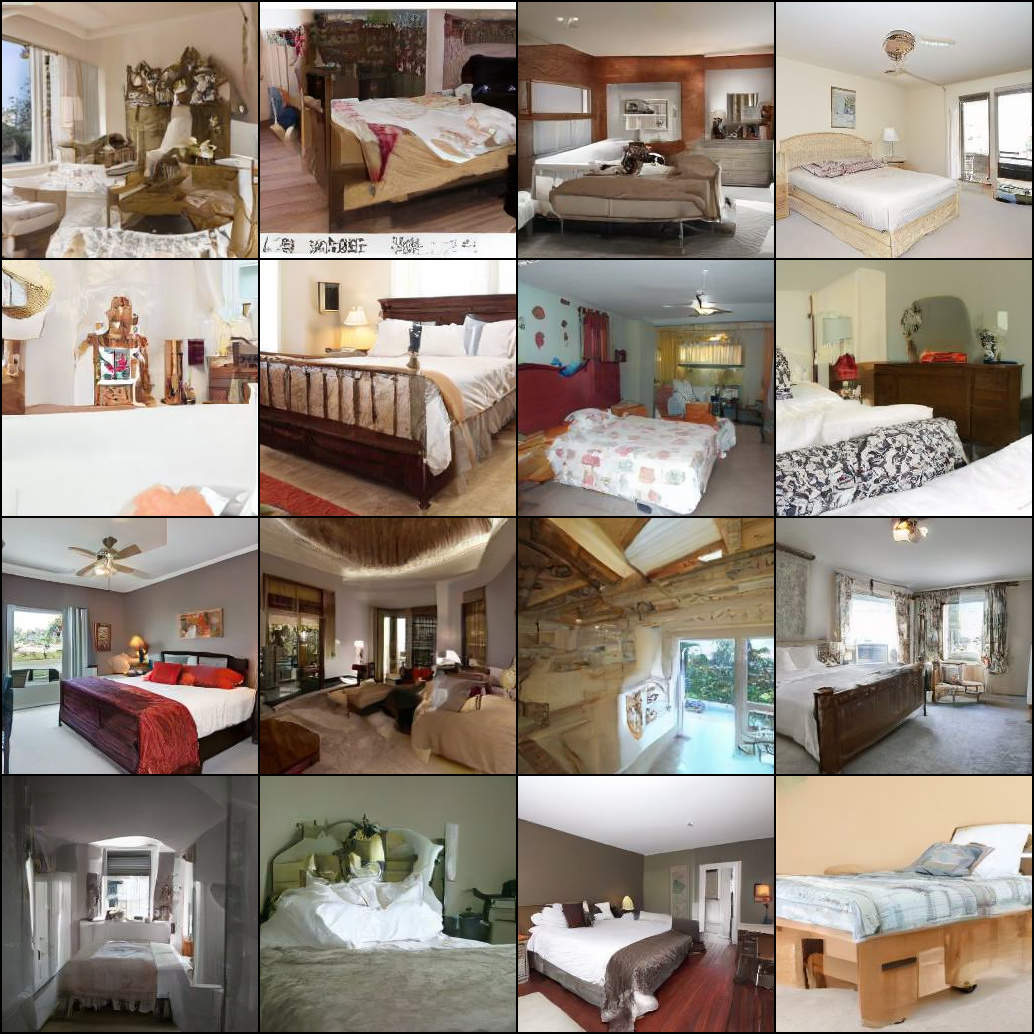}\label{fig:bedroom_b}
   }
\end{center}
   \caption{
     Visualization of generations on LSUN-Bedroom-256 (\textit{best viewed when zoomed-in}).
   }
   \vspace{-4mm}
   \label{fig:bedroom}
\end{figure}
\subsection{Qualitative Results}
\ourmodel is designed to directly facilitate the denoising task in wavelet sapce. As such, during sampling, we perform sequential reversed diffusion process to generate wavelets first.  
In Figure~\ref{fig:cifar_a},~\ref{fig:ffhq_a},~\ref{fig:church_a} and~\ref{fig:bedroom_a}, we generate high-quality wavelets with all four subbands information from noise (Top left: $\bU_{ll}$, Top right: $\bU_{lh}$, Bottom left: $\bU_{hl}$, Bottom right: $\bU_{hh}$).
These visualizations show that our model is able to (1) capture the spatial information to provide  realistic content in the synthesized image; and (2) approximate the complementary information from all four subbands to represent one image, including the highly semantic low-frequency content and the fine-grained details in high-frequency components.
The final generated images are obtained by performing inverse wavelet transform onto the wavelets, shown in Figure~\ref{fig:cifar_b},~\ref{fig:ffhq_b},~\ref{fig:church_b},~\ref{fig:bedroom_b}.

\begin{table*}[htb]
\footnotesize
\begin{center}
\tablestyle{6pt}{1.0} 
\caption{We ablate the effectiveness of different architectural components on FFHQ-256.}
\label{tab:ablation:result}
\begin{tabular}
{lccccccccc}
\toprule
&\multicolumn{1}{c}{{Methods}}
&\multicolumn{1}{c}{Spatial Conv.}
&\multicolumn{1}{c}{Frequency Conv.}
&\multicolumn{1}{c}{Spatial Attn.}
&\multicolumn{1}{c}{Frequency Attn.}
&\multicolumn{1}{c}{FID $\downarrow$} \\
\midrule
&Spatial-only Baseline &\cmark &\xmark &\cmark &\xmark &23.18\\
&+Freq-Conv. &\cmark &\cmark &\cmark  &\xmark &13.66\\
&+Freq-Attn.  &\cmark &\xmark &\cmark &\cmark &15.22\\
&\ourmodel &\cmark &\cmark &\cmark &\cmark  &\textbf{7.12}\\
\bottomrule
\end{tabular}
   \vspace{-5mm}
\end{center}
\end{table*}

\begin{figure}[htb]
\captionsetup[subfloat]{farskip=1pt,captionskip=1pt}
 \begin{center}
   \subfloat[Generated Wavelets]{
      \includegraphics[height = 0.13\columnwidth]{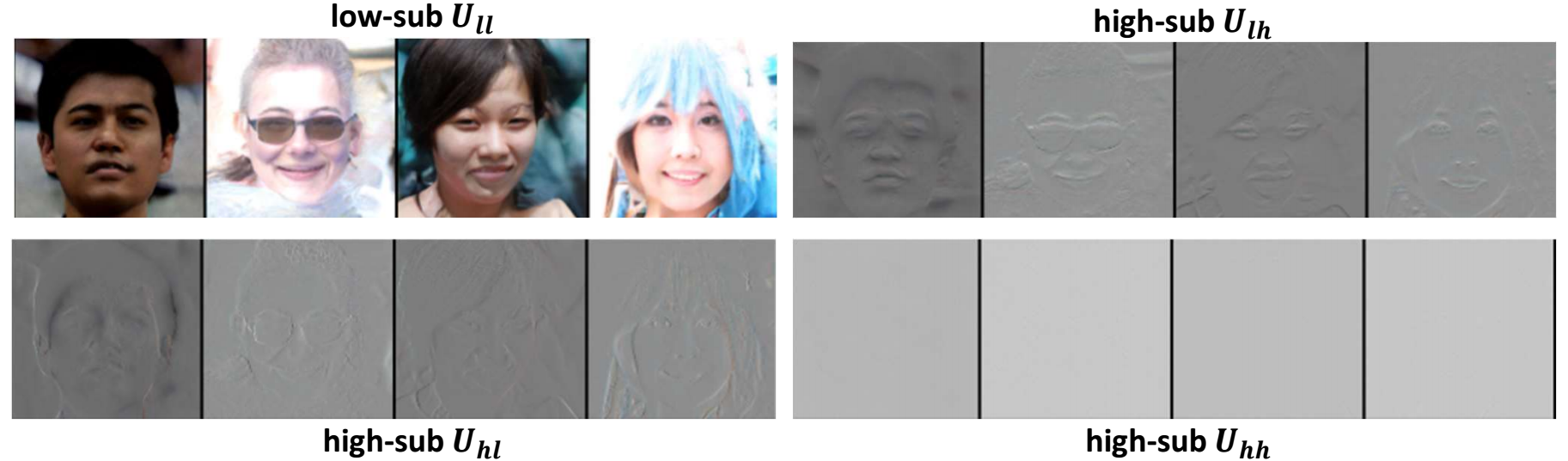}\label{fig:baseline_a}
   }\hfill
   \subfloat[Generated Images]{
      \includegraphics[height = 0.13\columnwidth]{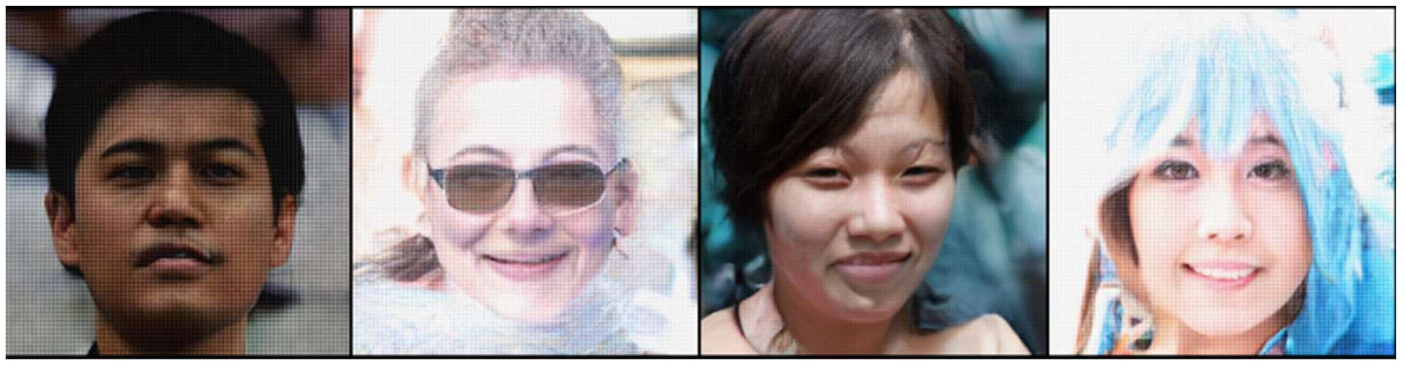}\label{fig:baseline_b}
   }
\end{center}
   \vspace{-1mm}
   \caption{
     FFHQ-256 samples generated by the spatial-only baseline model (1st row of Table~\ref{tab:ablation:result}).
   }
   \label{fig:baseline}
\end{figure}
\subsection{Ablation Study}
We ablate the effectiveness of the spatial-frequency convolutions and attentions in Table~\ref{tab:ablation:result} on FFHQ-256.
We first set up a simple baseline architecture with only spatial convolutions and attentions.
Then, we extend it to explicitly exploit the correlation along frequency dimension, by adding frequency convolutions and attentions, one at a time. The results show that all components need to work together to achieve the lowest FID score.
In addition, we visualize the generated wavelets and images by the spatial-only baseline architecture in Figure~\ref{fig:baseline}. As shown in the top left of Figure~\ref{fig:baseline_a}, the model with only spatial modules can still learn the low-frequency facial content (low-sub $\bU_{ll}$), however, it does not model the correlation among different frequency components, resulting in a noisy sample (especially for the high-subs) due to the failure of spatial-frequency alignment. Both quantitative and qualitative comparisons demonstrate that the proposed spatial-frequency-aware architecture is a necessary design for wavelet-based DDPMs.

\begin{figure}[htb]
\begin{center}
\centerline{\includegraphics[width=\columnwidth]{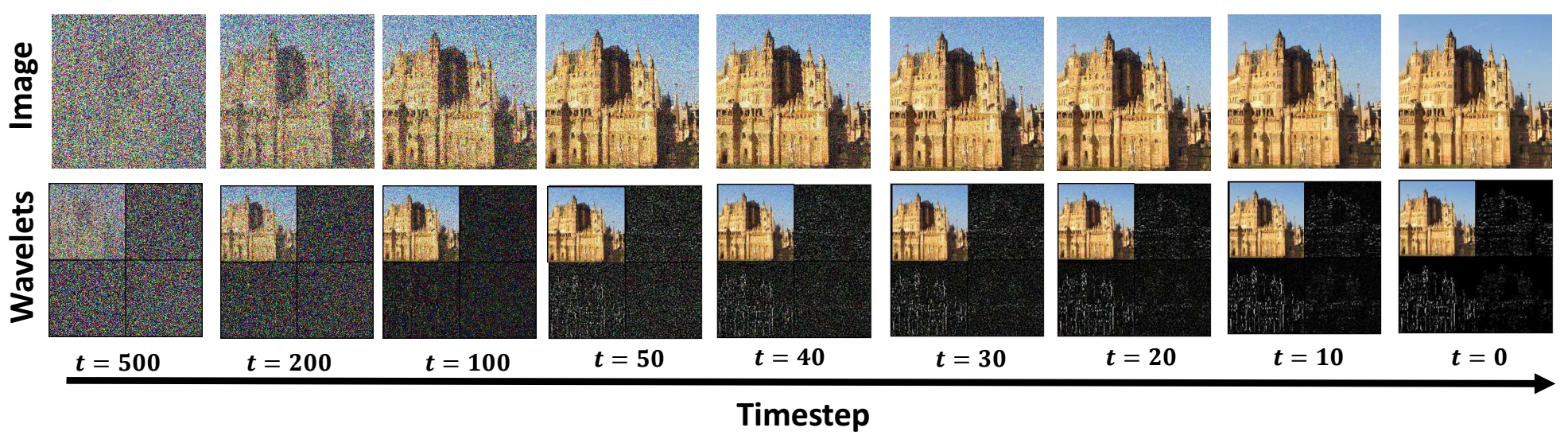}}
\caption{Illustration of wavelets and image refinement along the timesteps of reversed diffusion process.
}
\label{fig:wave_refine}
\vspace{-7mm}
\end{center}
\end{figure}
\subsection{Wavelets Refinement}
In Figure~\ref{fig:wave_refine}, we examine whether our method can explicitly recover the distribution of wavelets driven by the simple denoising objective, in contrast to the  reconstruction loss and the auxiliary adversarial loss in WaveDiff.
We demonstrate this by visualizing the predicted wavelets along the timesteps of reversed diffusion process. The generated images on the first row are from applying IWT to the generated wavelets at the corresponding timesteps.
We observe the wavelet generations in all four subbands get refined as the reversed diffusion process continues, in turn revealing more details in the generated images.

\section{Conclusion}
\textbf{Contributions.} \ourmodel is a new U-Net architecture for DDPMs in the wavelet domain that can produce realistic images in high quality.
With carefully designed spatial-frequency modules, \ourmodel obtains the capability in learning complementary information from  all frequency subspaces and capturing the spatial coherence simultaneously.
Though \ourmodel contains several components, it is not a collection of orthogonal innovations.
Rather, these components are designed to weave together for a high-level vision: enabling effective DDPM training in the wavelet space while maintaining the compatibility with the standard diffusion process and minimizing the efforts in designing a new optimization recipe.
Quantitative and qualitative results on multiple datasets, together with ablations and analyses demonstrate the effectiveness of our method.

\textbf{Limitations and Future Work.}
We investigated how to effectively adapt DDPM training process to input signals from wavelet space,
and proposed a new architecture that explicitly learns to recover all  frequency components guided by the simple denoising objective.
The limitation of this work sheds light on potential future directions, \textit{e.g.}, to conduct deeper analysis onto the generated high-frequency components, with the goal of formulating a more fine-grained image processing framework through controllable wavelets generation and editing.

\textbf{Broader Impacts.}
The improved image quality in DDPM with our proposed \ourmodel may bring up new possibilities in application scenarios, however,  it also
requires
proper regulations for mitigating potential harm from the misuse of such generative models, for example the creation of deceptive content and 
infringement of human rights.


\small
\bibliography{ref}
\bibliographystyle{plain}

\newpage
\appendix

\noindent \textbf{\Large Appendix}
\vspace{0.1in}

This appendix is organized as follows:
\begin{itemize}
    \item Section~\ref{supp_sec:3d} details the implementation and comparison for 3D-full baseline and \ourmodel on Bedroom.

    \item Section~\ref{supp_sec:channel_scale} ablates the model scalings for \ourmodel on FFHQ.
    
    \item Section~\ref{supp_sec:more} provides more qualitative and quantitative results at different sampling steps on Bedroom and Church.
\end{itemize}

\section{Comparison with 3D-full baseline}\label{supp_sec:3d}
We compare \ourmodel with a baseline model which uses 3D convolutions and attention at all spatial and frequency locations, denoted as 3D-full baseline.
\subsection{Implementations}
We implement 3D Convolution and 3D All Attention in Pytorch code~\ref{lst:3d} and~\ref{lst:3dattn}. We also provide the code implementations of our spat-freq conv and attention in~\ref{lst:21d} and~\ref{lst:sepattn}, respectively. Note that both methods share the consistent QKV self-attention implementation in code~\ref{lst:qkv}.

\textbf{Convolutions}
\renewcommand{\lstlistingname}{Code}
\begin{lstlisting}[language=Python, label={lst:3d}, caption=3D Convolution]
dims=3 #3D conv
ops = nn.Sequential(
        normalization(in_channels),
        SiLU(), #Nonlinear
        conv_nd(dims, in_channels, out_channels, stride=(1,1,1), 
        kernel_size=(3,3,3), padding=(1,1,1))
        )
\end{lstlisting}

\begin{lstlisting}[language=Python, label={lst:21d}, caption=(2+1)D Spat-Freq Convolution]
f = 3 #kernel_size for frequency
k = 3 #kernel_size for spatial
dims = 3

midplanes = (in_channels * out_channels * f * k * k) //
            (channels * k * k + f * out_channels)
ops = nn.Sequential(
        normalization(in_channels),
        SiLU(), #Nonlinear
        conv_nd(dims, in_channels, self.midplanes, stride=(1,1,1), 
        kernel_size=(1,3,3), padding=(0,1,1)), #2D spatial
        normalization(midplanes),
        SiLU(), #Nonlinear
        conv_nd(dims, midplanes, out_channels, stride=(1,1,1), 
        kernel_size=(3,1,1), padding=(1,0,0)) #1D frequency
        )
\end{lstlisting}

\noindent \textbf{Attentions}
\begin{lstlisting}[language=Python, label={lst:qkv},caption=QKV Self-Attention (red blcok in Figure 2 of the main paper)]
import torch
class QKVAttention(nn.Module):
    """
    A module which performs QKV attention.
    """

    def forward(self, qkv):
        """
        Apply QKV attention.

        :param qkv: an [N x (C * 3) x T] tensor of Qs, Ks, and Vs.
        :return: an [N x C x T] tensor after attention.
        """
        ch = qkv.shape[1] // 3
        q, k, v = torch.split(qkv, ch, dim=1)
        scale = 1 / math.sqrt(math.sqrt(ch))
        weight = torch.einsum(
            "bct,bcs->bts", q * scale, k * scale
        )  # More stable with f16 than dividing afterwards
        weight = torch.softmax(weight.float(), dim=-1).type(weight.dtype)
        return torch.einsum("bts,bcs->bct", weight, v)
\end{lstlisting}

\begin{lstlisting}[language=Python, label={lst:3dattn}, caption=All Attention (in 3D-full baseline)]
class 3DAllAttentionBlock(nn.Module):
    """
    An attention block that allows all (spatial and freqency) positions to attend to each other.
    """
    
    def forward(self, x):
        b, c, *spat_freq = x.shape
        x = x.reshape(b, c, -1)
        qkv = QKVAttention(self.norm(x)) #QKV Self-Attention
        qkv = qkv.reshape(b * self.num_heads, -1, qkv.shape[2])
        h = self.attention(qkv)
        h = h.reshape(b, -1, h.shape[-1])
        h = self.proj_out(h)
        return (x + h).reshape(b, c, *spat_freq)
\end{lstlisting}

\begin{lstlisting}[language=Python, label={lst:sepattn}, caption=Spatial-Frequency Attention]
class AttentionBlock(nn.Module):
    """
    An attention block that allows spatial-only positions to attend to each other.
    """
    
    def forward(self, x):
        b, c, f, s1, s2 = x.shape
        x = x.permute((0,2,1,3,4)).reshape(b*f, c, -1) #permutation
        qkv = QKVAttention(self.norm(x)) #QKV Self-Attention
        qkv = qkv.reshape(b * f * self.num_heads, -1, qkv.shape[2])
        h = self.attention(qkv)
        h = h.reshape(b*f, -1, h.shape[-1])
        h = self.proj_out(h)
        return (x + h).reshape(b, f, c, s1, s2).permute((0,2,1,3,4))

class AttentionBlockFreq(nn.Module):
    """
    An attention block that allows frequency-only positions to attend to each other.
    """
    
    def forward(self, x):
        b, c, f, s1, s2 = x.shape
        x = x.permute((0,3,4,1,2)).reshape(b*s1*s2, c, -1) #permutation
        qkv = QKVAttention(self.norm(x)) #QKV Self-Attention
        qkv = qkv.reshape(b * s1 *s2 * self.num_heads, -1, qkv.shape[2])
        h = self.attention(qkv)
        h = h.reshape(b*s1*s2, -1, h.shape[-1])
        h = self.proj_out(h)
        return (x + h).reshape(b, s1, s2, c, f).permute((0,3,4,1,2))
\end{lstlisting}

\begin{figure}[tbh]
\captionsetup[subfloat]{farskip=1pt,captionskip=1pt}
   \begin{center}
      \includegraphics[width=\columnwidth]{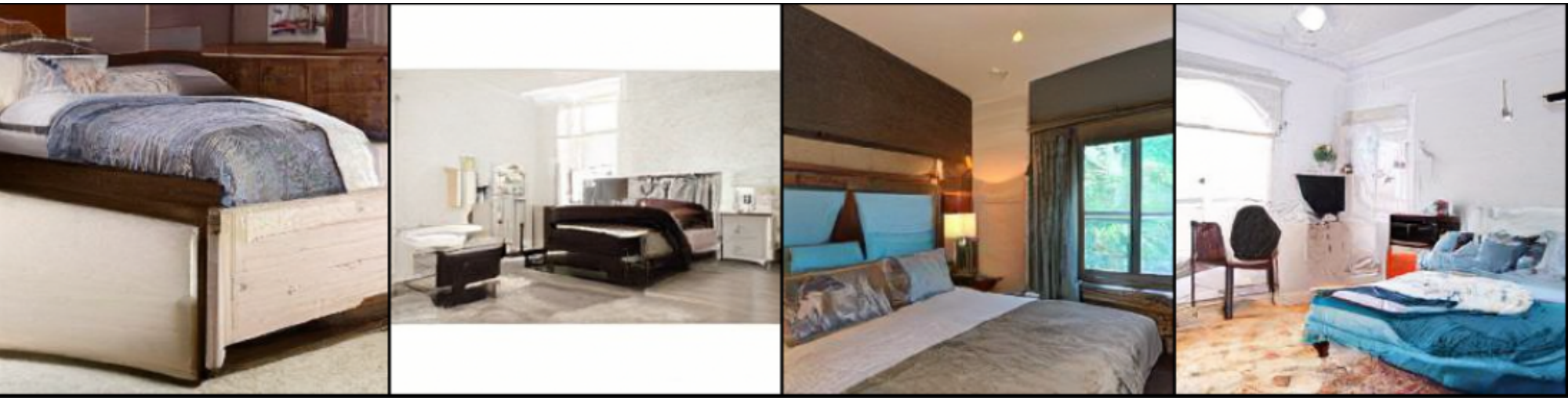}
   \end{center}
   \caption{Generation Results on LSUN-Bedroom using 3D-full baseline.
   }
   \label{append_fig:bedroom:3d}
\end{figure}

\textbf{Results}
As shown in Figure~\ref{append_fig:bedroom:3d}, 3D-full baseline is able to generate realistic images with high-quality details. 
This is because 3D convolutions and attentions at all locations altogether can also recover all frequency sub-bands with complementary information. 
Nevertheless, Table~\ref{tab:full3d:result} suggests that the spat-freq design in \ourmodel is more efficient and effective.

\begin{table}[tbh]
\begin{center}
\caption{Qunatitative comparison between 3D-full and \ourmodel on LSUN-Bedroom.}
\label{tab:full3d:result}
\begin{tabular}
{lccccccccc}
\toprule
&\multicolumn{1}{c}{{Methods}}
&\multicolumn{1}{c}{{Params (M)}}
&\multicolumn{1}{c}{{FLOPs (G)}}
&\multicolumn{1}{c}{{FID}} \\
\midrule
& 3D-full &364.14 &870.06 &6.73\\
& \ourmodel &291.31 &669.28 &3.88\\
\bottomrule
\end{tabular}
\end{center}
\end{table}

\begin{table}[tbh]
\begin{center}
\vspace{-4mm}
\caption{Scaling effect with model size on FFHQ}
\label{tab:scale:result}
\begin{tabular}
{lccccccccc}
\toprule
&\multicolumn{1}{c}{{Model Scale}}
&\multicolumn{1}{c}{{Params (M)}}
&\multicolumn{1}{c}{{FLOPs (G)}}
&\multicolumn{1}{c}{{FID}} \\
\midrule
& $c=64$ &73.10 &163.01 &43.12 \\
& $c=128$ (default) &291.31 &669.28 &12.48 \\
& $c=192$ &665.35 &1496.23 &10.09\\
\bottomrule
\end{tabular}
\end{center}
\vspace{-8mm}
\end{table}

\begin{figure}[tbh]
\captionsetup[subfloat]{farskip=1pt,captionskip=1pt}
   \begin{center}
   \subfloat[$c=64$, 73M params]{
      \includegraphics[height = 0.3 \columnwidth]{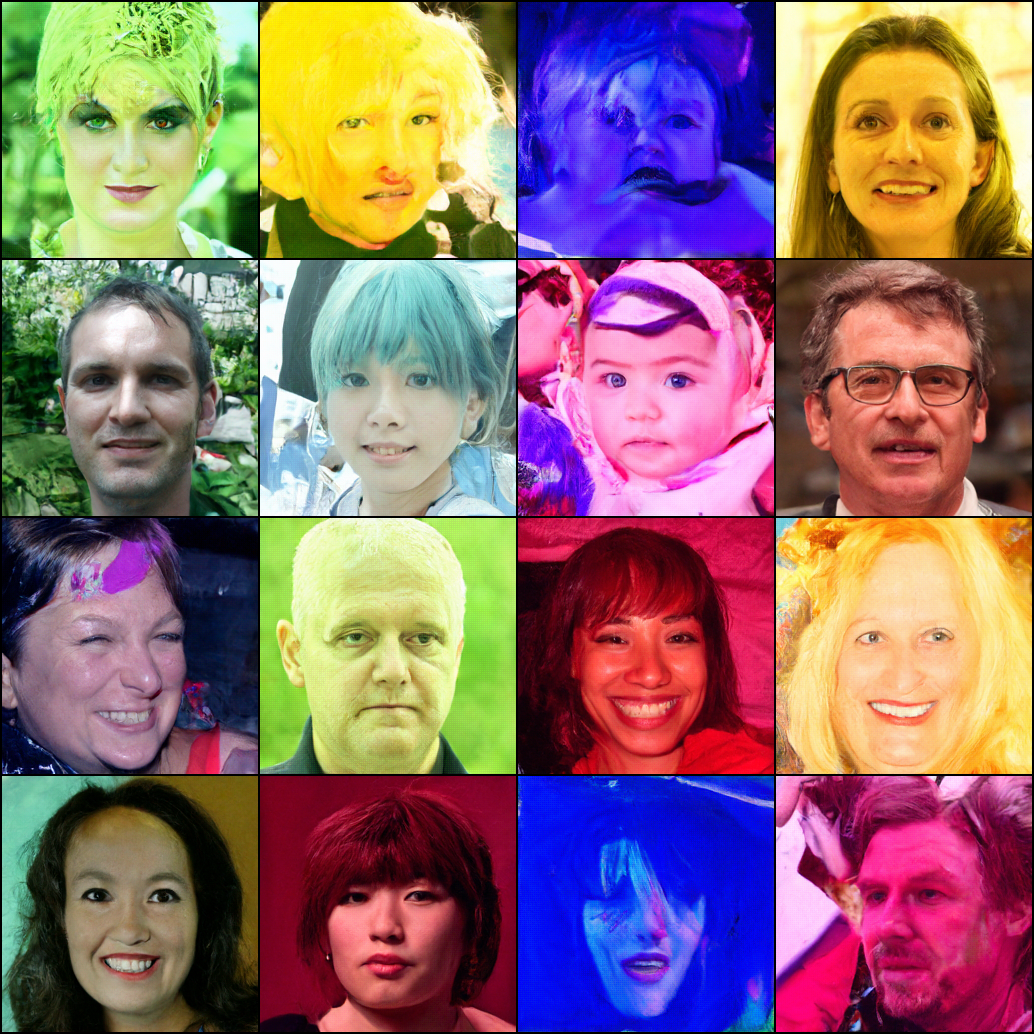}
   }\hfill
   \subfloat[$c=128$, 291M params]{
      \includegraphics[height = 0.3 \columnwidth]{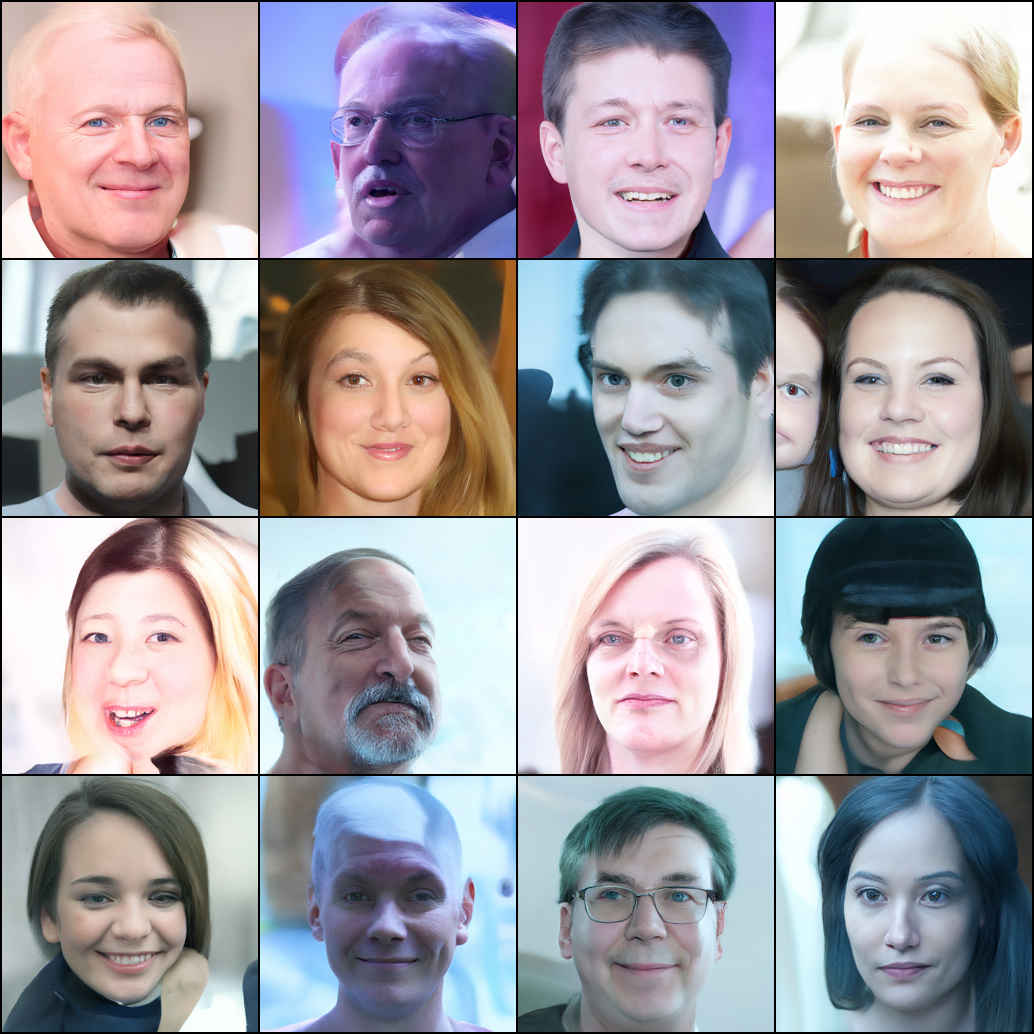}
   }
   \hfill
   \subfloat[$c=192$, 665M params]{
      \includegraphics[height = 0.3 \columnwidth]{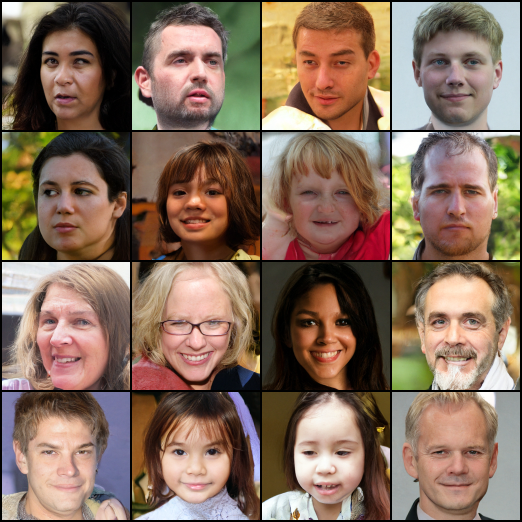}
   }
   \end{center}

   \caption{FFHQ generation using model of different sizes (train 100K iterations)
   }
   \vspace{-4mm}
   \label{append_fig:ffhq:size}
\end{figure}

\section{Scaling Model Size}\label{supp_sec:channel_scale}
To measure how performance scales with model size, we train another two models for 100K iterations on FFHQ, with different base channels $c=64, 192$, respectively. The results in Table~\ref{append_fig:ffhq:size} show that the sample quality improves as model size and computation increases. However, naively scaling the model size may not be the optimal solution to better generation performance. As observed in Table~\ref{append_fig:ffhq:size}, when scaling the model from 291M to more than 600M parameters, the performance improvement becomes more subtle (-2.39 in FID) compared to scaling from 73M to 291M parameters (-30.64 in FID). 

\section{Generation with Reduced Sampling Steps}\label{supp_sec:more}
\begin{figure}[tbh]
\captionsetup[subfloat]{farskip=1pt,captionskip=1pt}
 \begin{center}
   \subfloat[Generated Images]{
      \includegraphics[width = 0.48 \columnwidth]{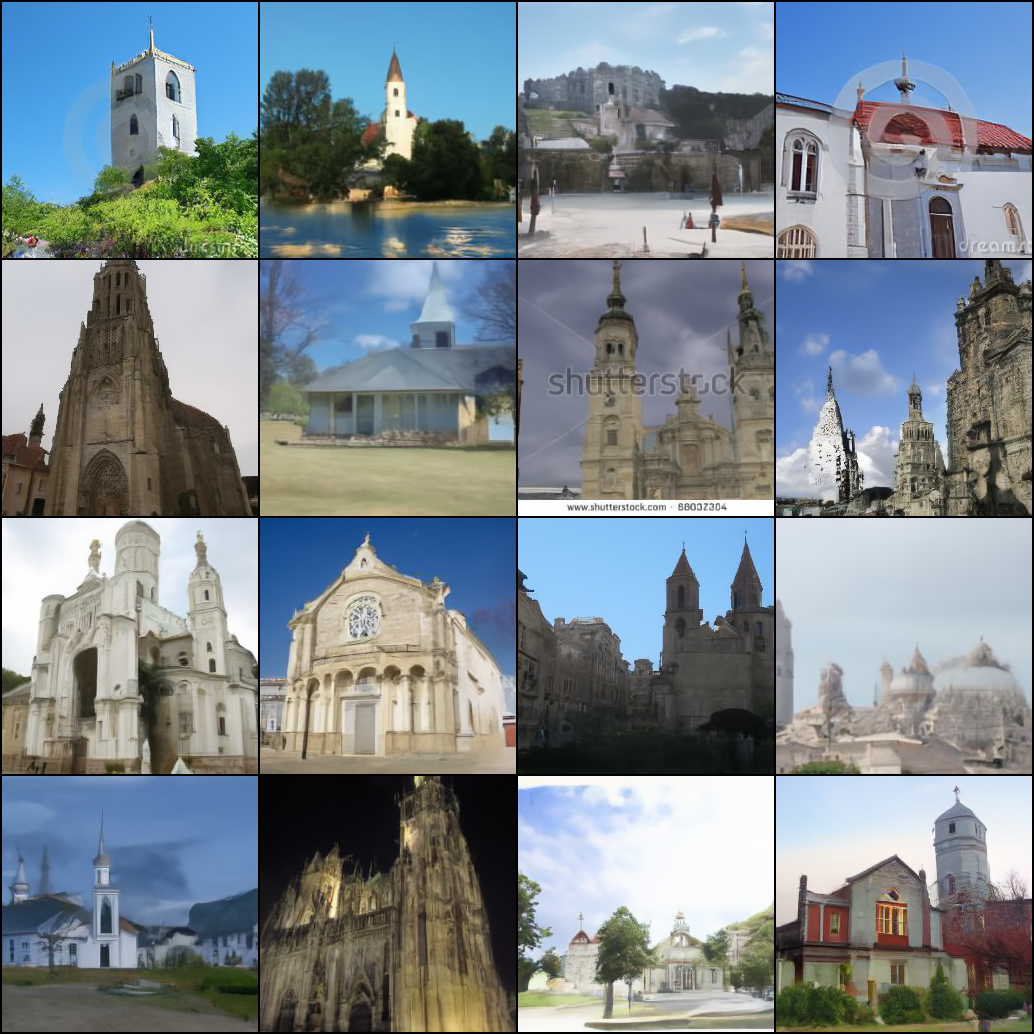}
   }
   \end{center}
   \begin{center}
   \subfloat[Gen. Wavelets $\bU_{ll}$ ]{
      \includegraphics[height = 0.22 \columnwidth]{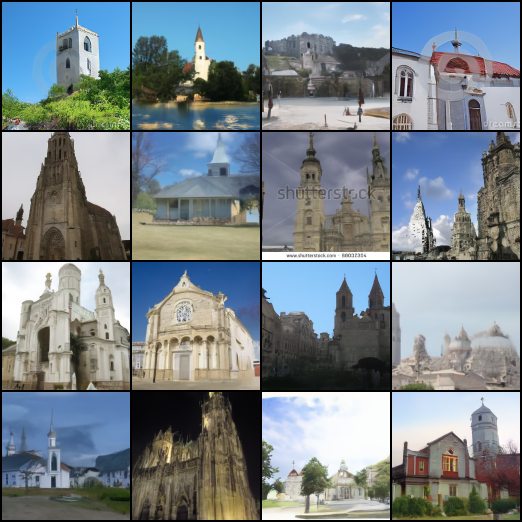}
   }\hfill
   \subfloat[Gen. Wavelets $\bU_{lh}$  ]{
      \includegraphics[height = 0.22 \columnwidth]{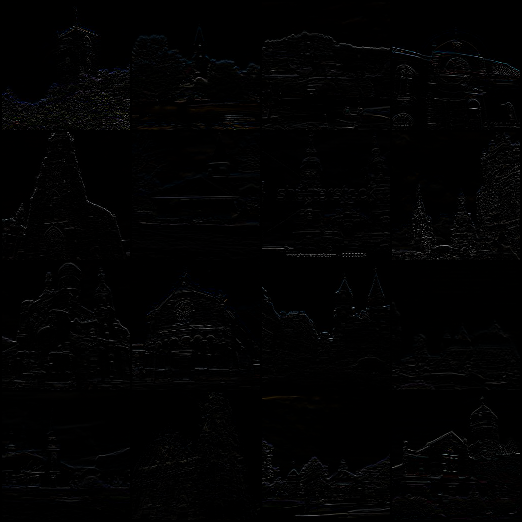}
   }\hfill
   \subfloat[Gen. Wavelets $\bU_{hl}$ ]{
      \includegraphics[height = 0.22 \columnwidth]{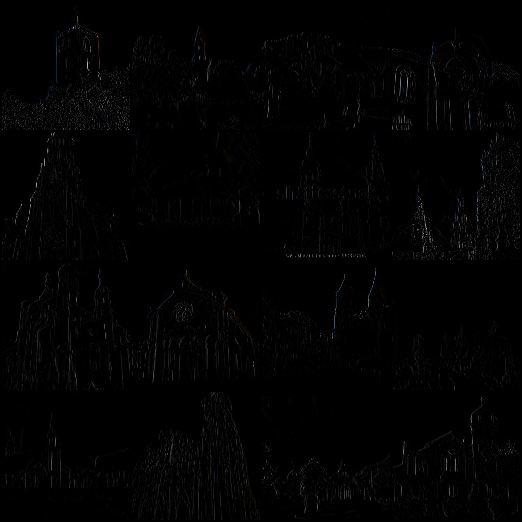}
   }\hfill
   \subfloat[Gen. Wavelets $\bU_{hh}$ ]{
      \includegraphics[height = 0.22 \columnwidth]{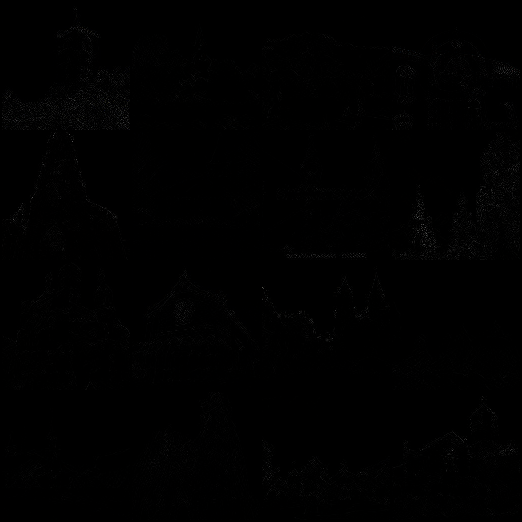}
   }
   \end{center}

   \caption{LSUN-Church generation @50 sampling steps.
   }
   \vspace{-4mm}
   \label{append_fig:church:50}
\end{figure}

\begin{figure}[tbh]
\captionsetup[subfloat]{farskip=1pt,captionskip=1pt}
 \begin{center}
   \subfloat[Generated Images]{
      \includegraphics[width = 0.48 \columnwidth]{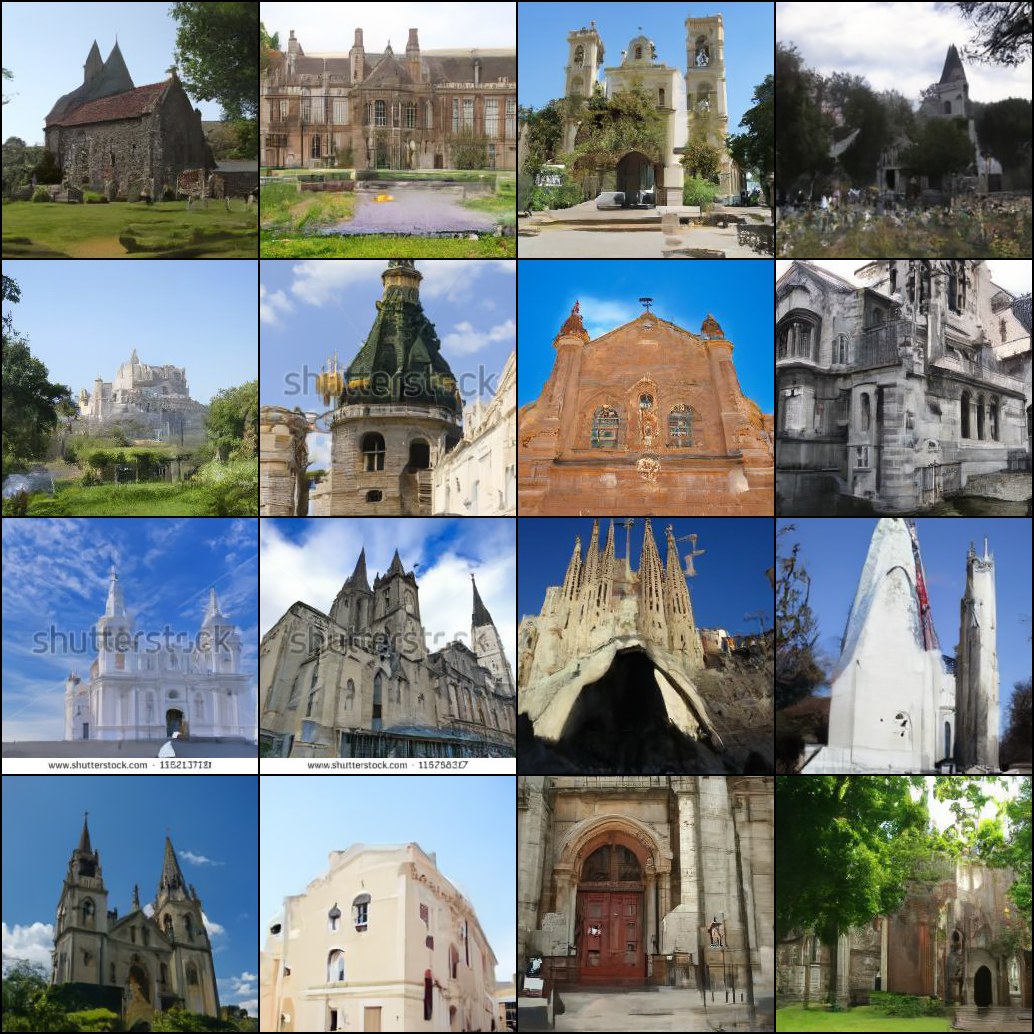}
   }
   \end{center}
   \begin{center}
   \subfloat[Gen. Wavelets $\bU_{ll}$ ]{
      \includegraphics[height = 0.22 \columnwidth]{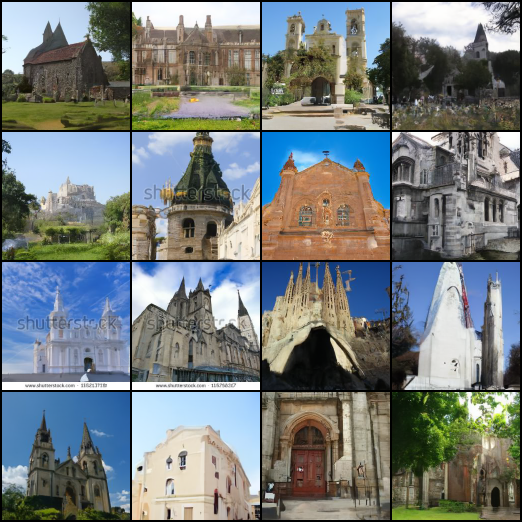}
   }\hfill
   \subfloat[Gen. Wavelets $\bU_{lh}$  ]{
      \includegraphics[height = 0.22 \columnwidth]{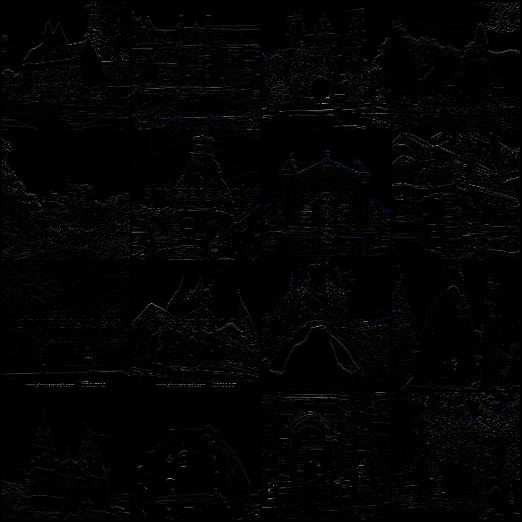}
   }\hfill
   \subfloat[Gen. Wavelets $\bU_{hl}$ ]{
      \includegraphics[height = 0.22 \columnwidth]{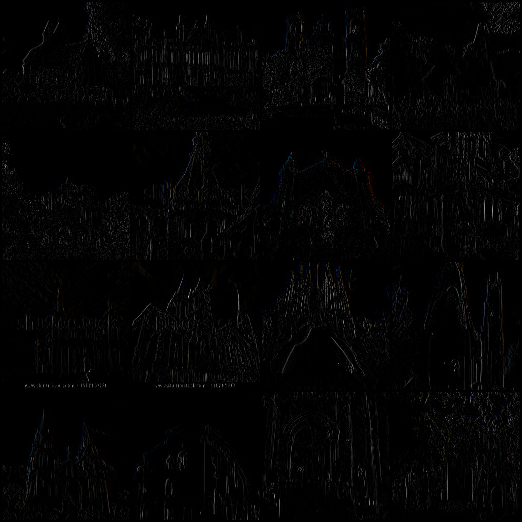}
   }\hfill
   \subfloat[Gen. Wavelets $\bU_{hh}$ ]{
      \includegraphics[height = 0.22 \columnwidth]{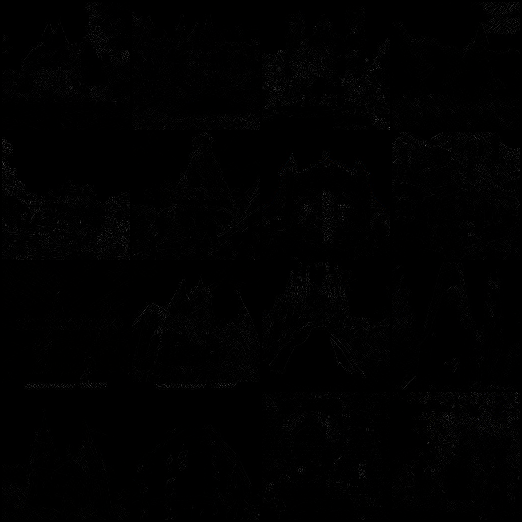}
   }
   \end{center}

   \caption{LSUN-Church generation @100 sampling steps.
   }
   \vspace{-4mm}
   \label{append_fig:church:100}
\end{figure}

\begin{figure}[tbh]
\captionsetup[subfloat]{farskip=1pt,captionskip=1pt}
 \begin{center}
   \subfloat[Generated Images]{
      \includegraphics[width = 0.48 \columnwidth]{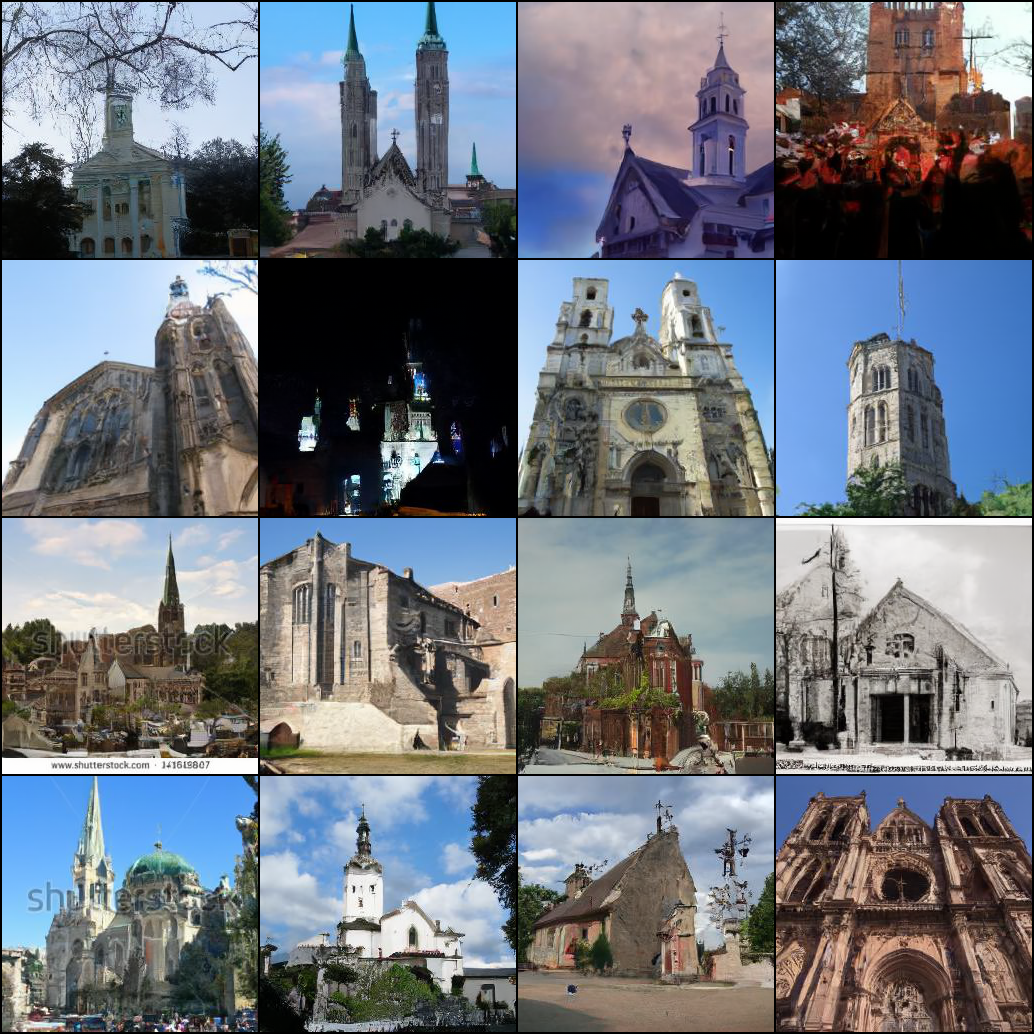}
   }
   \end{center}
   \begin{center}
   \subfloat[Gen. Wavelets $\bU_{ll}$ ]{
      \includegraphics[height = 0.22 \columnwidth]{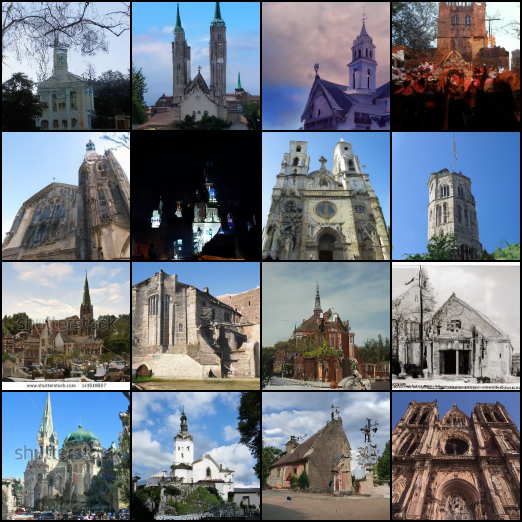}
   }\hfill
   \subfloat[Gen. Wavelets $\bU_{lh}$  ]{
      \includegraphics[height = 0.22 \columnwidth]{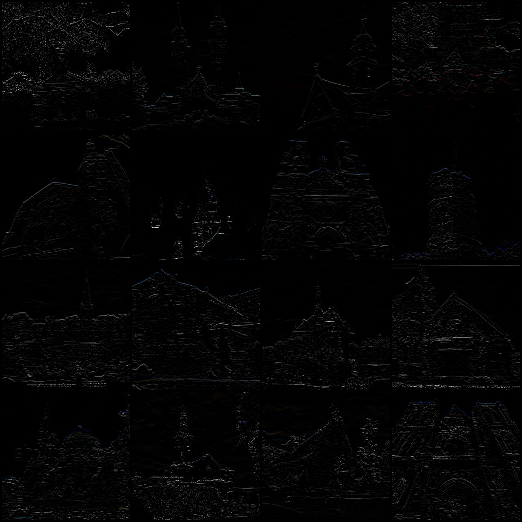}
   }\hfill
   \subfloat[Gen. Wavelets $\bU_{hl}$ ]{
      \includegraphics[height = 0.22 \columnwidth]{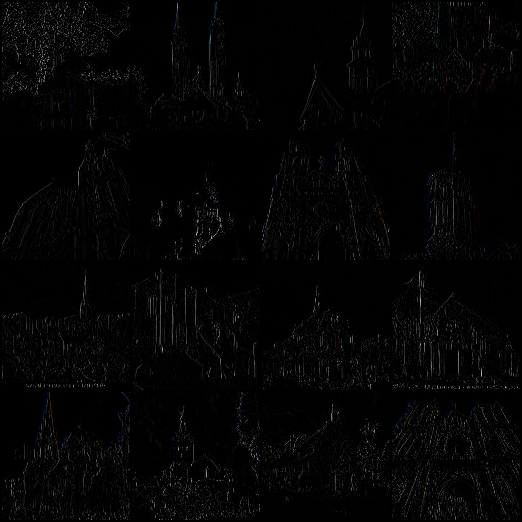}
   }\hfill
   \subfloat[Gen. Wavelets $\bU_{hh}$ ]{
      \includegraphics[height = 0.22 \columnwidth]{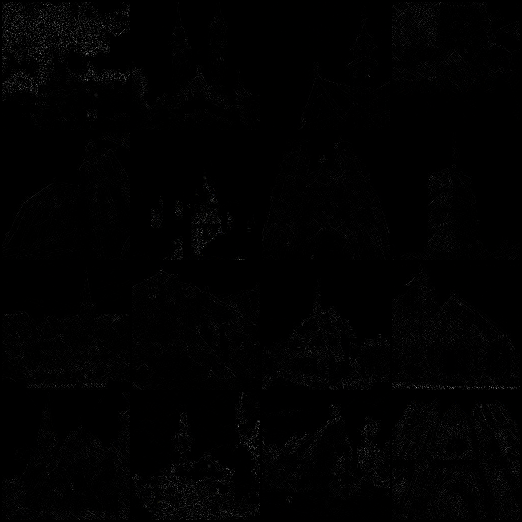}
   }
   \end{center}

   \caption{LSUN-Church generation @250 sampling steps.
   }
   \vspace{-4mm}
   \label{append_fig:church:250}
\end{figure}

\begin{figure}[tbh]
\captionsetup[subfloat]{farskip=1pt,captionskip=1pt}
 \begin{center}
   \subfloat[Generated Images]{
      \includegraphics[width = 0.48 \columnwidth]{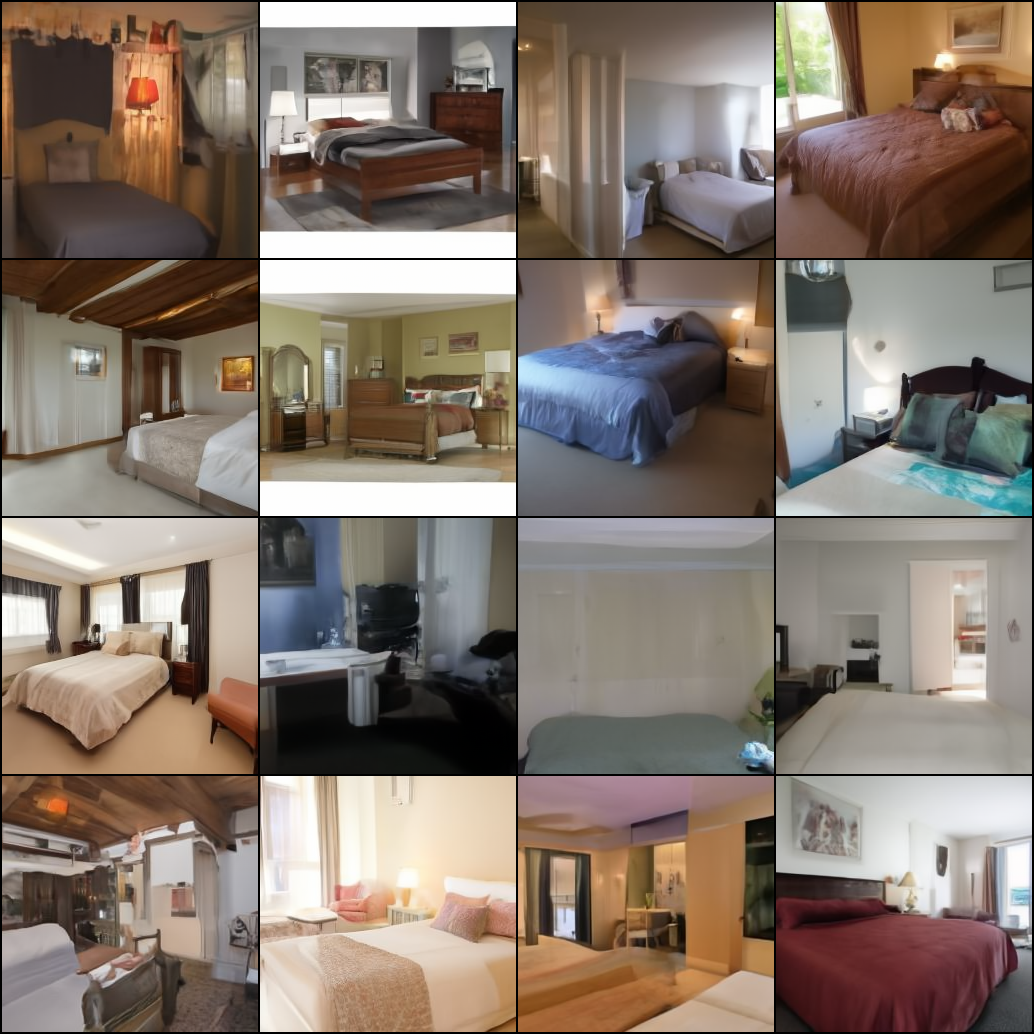}
   }
   \end{center}
   \begin{center}
   \subfloat[Gen. Wavelets $\bU_{ll}$ ]{
      \includegraphics[height = 0.22 \columnwidth]{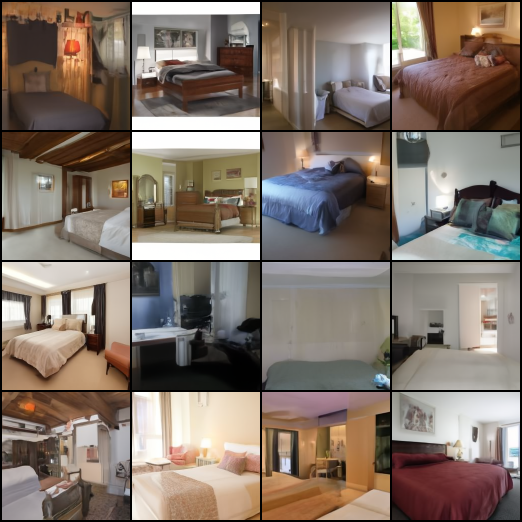}
   }\hfill
   \subfloat[Gen. Wavelets $\bU_{lh}$  ]{
      \includegraphics[height = 0.22 \columnwidth]{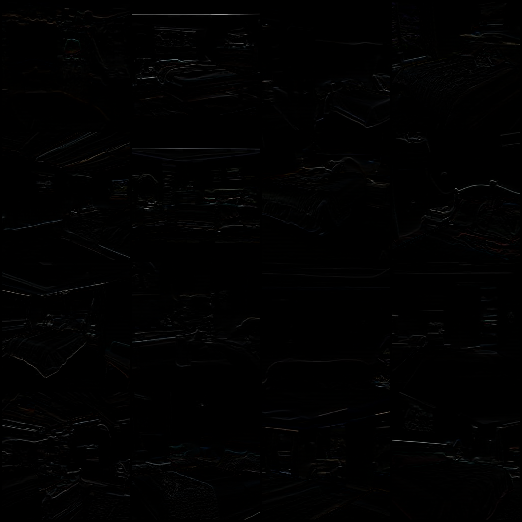}
   }\hfill
   \subfloat[Gen. Wavelets $\bU_{hl}$ ]{
      \includegraphics[height = 0.22 \columnwidth]{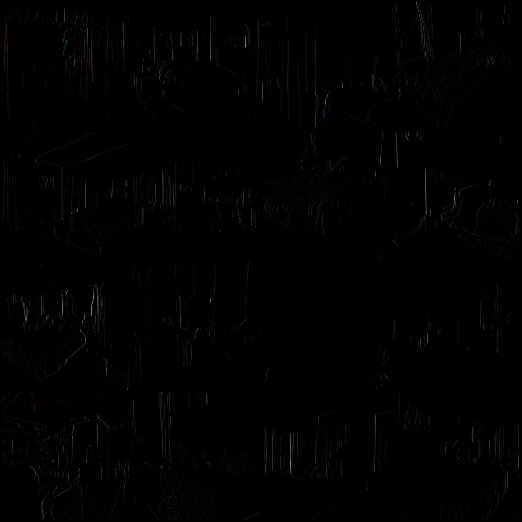}
   }\hfill
   \subfloat[Gen. Wavelets $\bU_{hh}$ ]{
      \includegraphics[height = 0.22 \columnwidth]{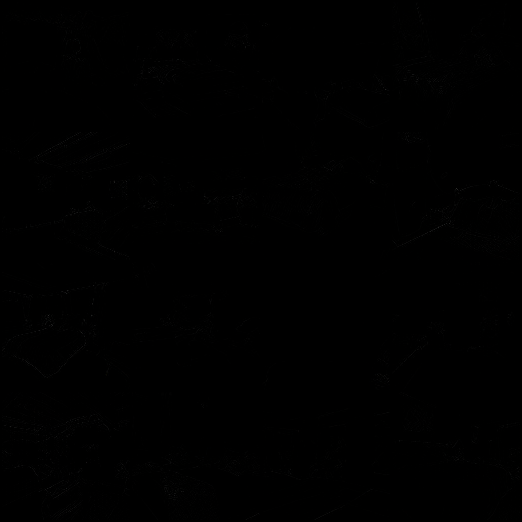}
   }
   \end{center}

   \caption{LSUN-Bedroom generation @50 sampling steps.
   }
   \vspace{-4mm}
   \label{append_fig:bedroom:50}
\end{figure}

\begin{figure}[tbh]
\captionsetup[subfloat]{farskip=1pt,captionskip=1pt}
 \begin{center}
   \subfloat[Generated Images]{
      \includegraphics[width = 0.48 \columnwidth]{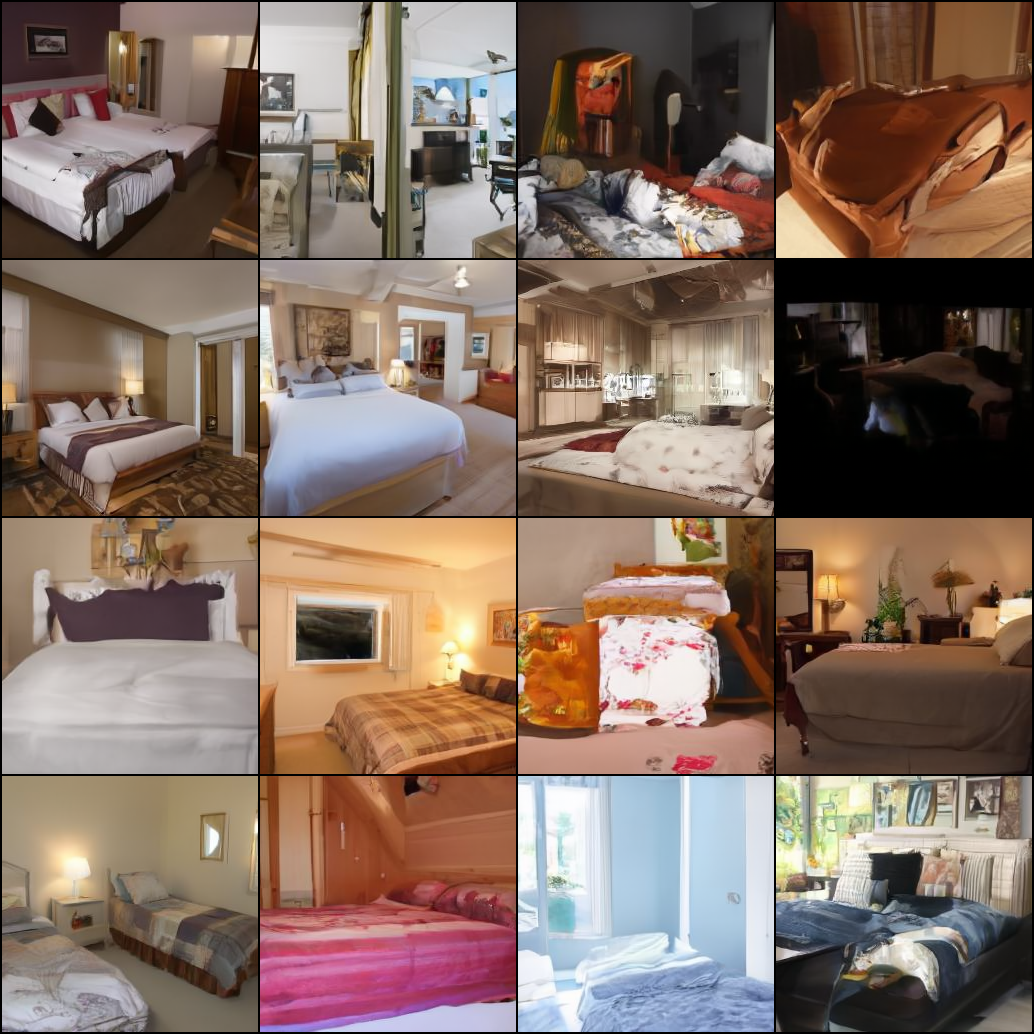}
   }
   \end{center}
   \begin{center}
   \subfloat[Gen. Wavelets $\bU_{ll}$ ]{
      \includegraphics[height = 0.22 \columnwidth]{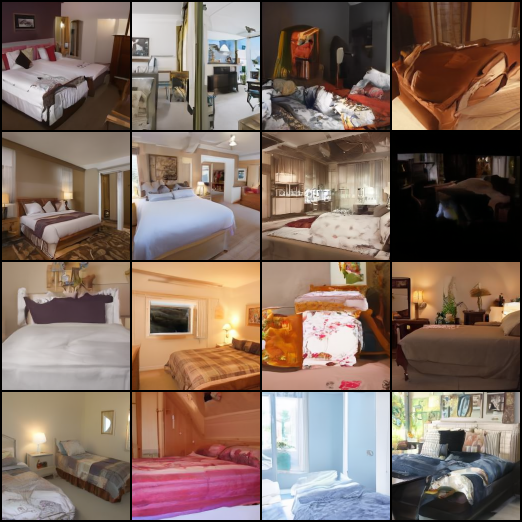}
   }\hfill
   \subfloat[Gen. Wavelets $\bU_{lh}$  ]{
      \includegraphics[height = 0.22 \columnwidth]{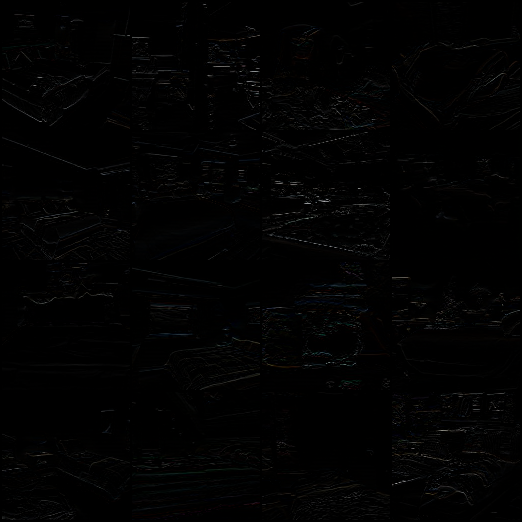}
   }\hfill
   \subfloat[Gen. Wavelets $\bU_{hl}$ ]{
      \includegraphics[height = 0.22 \columnwidth]{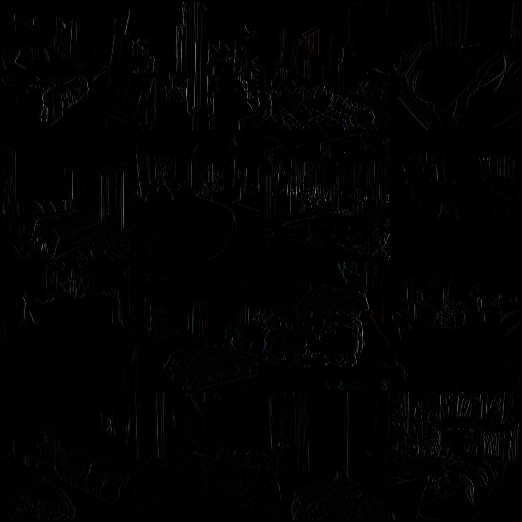}
   }\hfill
   \subfloat[Gen. Wavelets $\bU_{hh}$ ]{
      \includegraphics[height = 0.22 \columnwidth]{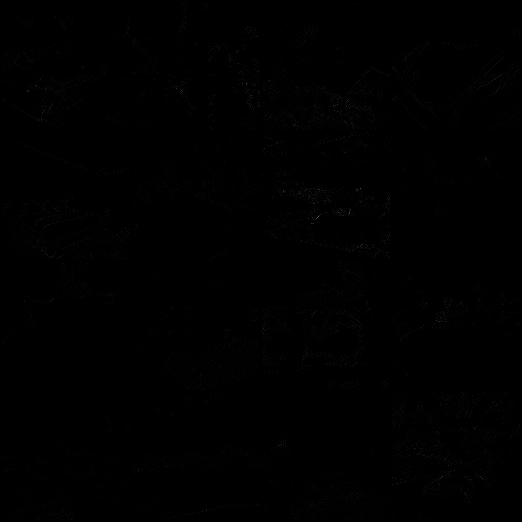}
   }
   \end{center}

   \caption{LSUN-Bedroom generation @100 sampling steps.
   }
   \vspace{-4mm}
   \label{append_fig:bedroom:100}
\end{figure}

\begin{figure}[tbh]
\captionsetup[subfloat]{farskip=1pt,captionskip=1pt}
 \begin{center}
   \subfloat[Generated Images]{
      \includegraphics[width = 0.48 \columnwidth]{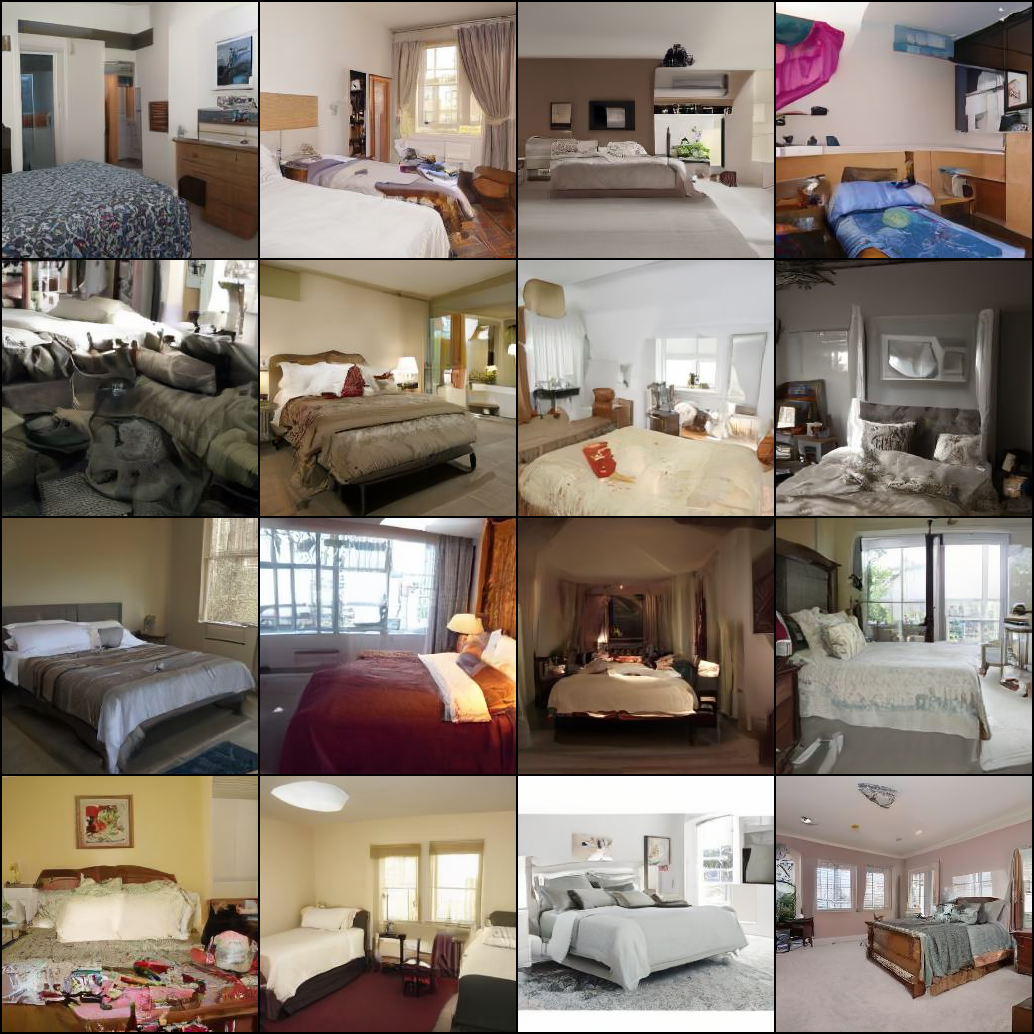}
   }
   \end{center}
   \begin{center}
   \subfloat[Gen. Wavelets $\bU_{ll}$ ]{
      \includegraphics[height = 0.22 \columnwidth]{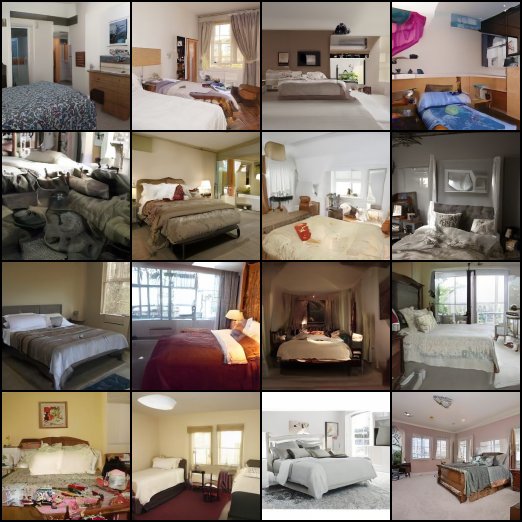}
   }\hfill
   \subfloat[Gen. Wavelets $\bU_{lh}$  ]{
      \includegraphics[height = 0.22 \columnwidth]{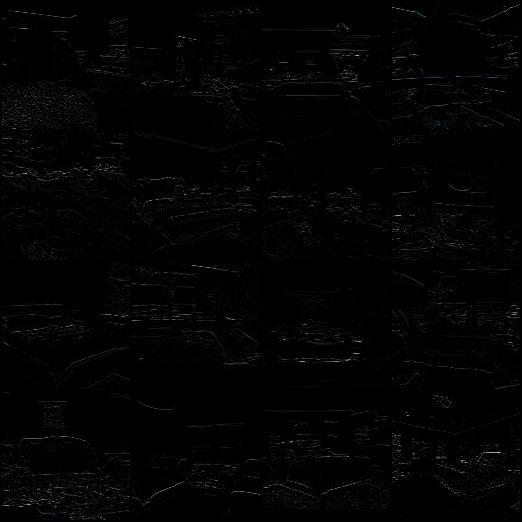}
   }\hfill
   \subfloat[Gen. Wavelets $\bU_{hl}$ ]{
      \includegraphics[height = 0.22 \columnwidth]{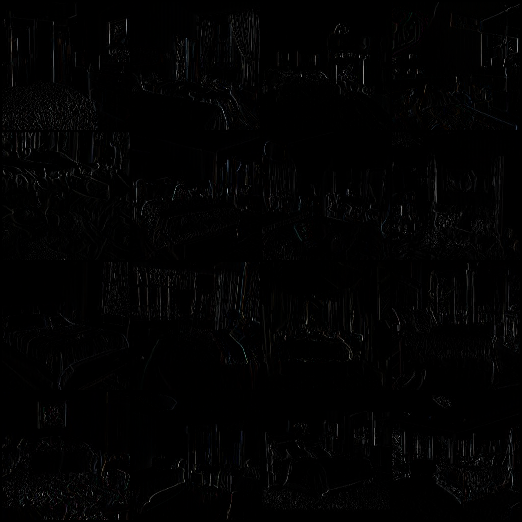}
   }\hfill
   \subfloat[Gen. Wavelets $\bU_{hh}$ ]{
      \includegraphics[height = 0.22 \columnwidth]{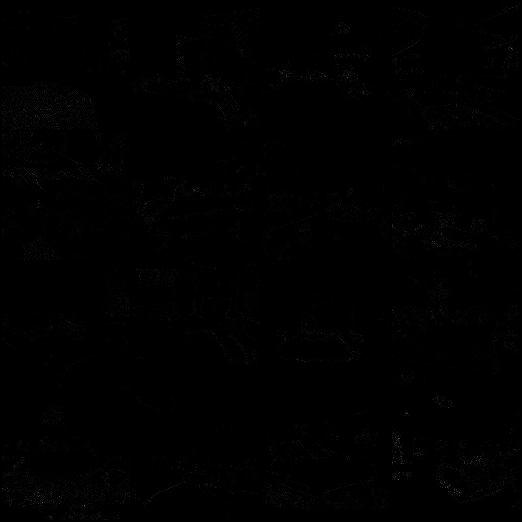}
   }
   \end{center}

   \caption{LSUN-Bedroom generation @250 sampling steps.
   }
   \vspace{-4mm}
   \label{append_fig:bedroom:250}
\end{figure}

We evaluate the models that were trained with 1000 sampling steps on Bedroom and Church using 50, 100, 150, 200 and 250 sampling steps during inference. As shown in Table~\ref{tab:sampling:result}, sampling with 1000 steps achieves the best FID. We also visualize the generation @50, 100, 250 sampling steps in Figure~\ref{append_fig:church:50}-~\ref{append_fig:bedroom:250} (zoom in for better view, especially for high-frequency sub-bands). We see even at 50 sampling steps, our method is able to recover the details of all frequency sub-bands hence producing realistic images. Sampling with 100 steps yields images with comparable visual quality.
\begin{table}[tbh]
\begin{center}
\vspace{-4mm}
\caption{FID($\downarrow$) with different sampling steps.}
\label{tab:sampling:result}
\begin{tabular}
{lccccccccc}
\toprule
&\multicolumn{1}{c}{{Dataset}}
&\multicolumn{1}{c}{{@50}}
&\multicolumn{1}{c}{{@100}}
&\multicolumn{1}{c}{{@150}}
&\multicolumn{1}{c}{{@200}}
&\multicolumn{1}{c}{{@250}}
&\multicolumn{1}{c}{{@1000}} \\
\midrule
&Church &10.65 &8.73 &8.00 &7.65 &7.00 &6.11\\
&Bedroom &11.23 &9.12 &6.22 &5.34 &4.77 &3.88\\
\bottomrule
\end{tabular}
\end{center}
\vspace{-8mm}
\end{table}



\end{document}